\documentclass[sigconf]{acmart}

\usepackage{enumitem}
\usepackage{xcolor,colortbl}
\usepackage{color, colortbl}
\usepackage{tabularx}
\usepackage{soul}

\newcommand{\rev}[1]{\textcolor{black}{#1}}

\newcommand{\Sref}[1]{\S\ref{#1}}

\definecolor{lightBlue}{rgb}{0.78, 0.85, 1.0}
\definecolor{lightOrange}{rgb}{0.88, 0.95, 1.0}
\definecolor{lightRed}{rgb}{1.0, 0.85, 0.85}
\usepackage{tcolorbox}
\usepackage{multirow}

\newtcbox{\bluebox}{on line, box align=base, colback=lightBlue,colframe=white,size=fbox,arc=3pt, before upper=\strut, top=-2pt, bottom=-4pt, left=-2pt, right=-2pt, boxrule=0pt}

\newtcbox{\orangebox}{on line, box align=base, colback=lightOrange,colframe=white,size=fbox,arc=3pt, before upper=\strut, top=-2pt, bottom=-4pt, left=-2pt, right=-2pt, boxrule=0pt}

\newtcbox{\redbox}{on line, box align=base, colback=lightRed,colframe=white,size=fbox,arc=3pt, before upper=\strut, top=-2pt, bottom=-4pt, left=-2pt, right=-2pt, boxrule=0pt}

\newcommand{\dashifted}{\raisebox{0.5\depth}{\tiny$\downarrow$}}
\newcommand{\upshifted}{\raisebox{0.5\depth}{\tiny$\uparrow$}}

\newcommand{\dar}[1]{{\small\redbox{\dashifted{#1}}}}

\newcommand{\uab}[1]{{\small\bluebox{\upshifted{#1}}}}

\newcommand{\titleFullNameFirst}{Call of Duty\textregistered: Modern Warfare\textregistered II}
\newcommand{\titleFullName}{Call of Duty: Modern Warfare II}

\newcommand{\company}{ABK}

\AtBeginDocument{%
  \providecommand\BibTeX{{%
    \normalfont B\kern-0.5em{\scshape i\kern-0.25em b}\kern-0.8em\TeX}}}

\setcopyright{acmlicensed}
\copyrightyear{2018}
\acmYear{2018}
\acmDOI{XXXXXXX.XXXXXXX}

\acmConference[Conference acronym 'XX]{Make sure to enter the correct
  conference title from your rights confirmation emai}{June 03--05,
  2018}{Woodstock, NY}
\acmISBN{978-1-4503-XXXX-X/18/06}

\begin{document}

\title{Self-Anchored Attention Model for Sample-Efficient Classification of Prosocial Text Chat}

\author{Zhuofang Li}
\authornote{Both authors contributed equally to this research.}
\email{zhuofang@caltech.edu}
\author{Rafal Kocielnik}
\authornotemark[1]
\email{rafalko@caltech.edu}
\affiliation{%
  \institution{California Institute of Technology}
  \city{Pasadena}
  \country{USA}
}

\author{Fereshteh Soltani}

\author{Penphob (Andrea) Boonyarungsrit}
\affiliation{%
  \institution{Activision-Blizzard-King}
  \city{Santa Monica}
  \country{USA}}
\email{feri.soltani@activision.com}
\email{andrea.boonyarungsrit@activision.com}

\author{Animashree Anandkumar}
\email{anima@caltech.edu}

\author{R. Michael Alvarez}
\email{rma@caltech.edu}
\affiliation{%
  \institution{California Institute of Technology}
  \city{Pasadena}
  \country{USA}
}

\renewcommand{\shortauthors}{Trovato and Tobin, et al.}

\begin{abstract}
Millions of players engage daily in competitive online games, communicating through in-game chat. Prior research has focused on detecting relatively small volumes of toxic content using various Natural Language Processing (NLP) techniques for the purpose of moderation. However, recent studies emphasize the importance of detecting prosocial communication, which can be as crucial as identifying toxic interactions. Recognizing prosocial behavior allows for its analysis, rewarding, and promotion. Unlike toxicity, there are limited datasets, models, and resources for identifying prosocial behaviors in game-chat text. In this work, we employed unsupervised \rev{discovery} combined with game domain expert collaboration to identify and categorize prosocial player behaviors from game chat. We further propose a novel \emph{Self-Anchored Attention Model (SAAM)} which gives \rev{7.9\%} improvement compared to the best existing technique. The approach utilizes the entire training set as \emph{``anchors''} to help improve model performance under the scarcity of training data. This approach led to the development of the first automated system for classifying prosocial behaviors in in-game chats, particularly given the low-resource settings where large\rev{-scale} labeled \rev{data is} not available. Our methodology was applied to one of the most popular online gaming titles - \titleFullNameFirst{}, showcasing its effectiveness. This research is novel in applying NLP techniques to discover and classify prosocial behaviors in player in-game chat communication. It can help shift the focus of moderation from solely penalizing toxicity to actively encouraging positive interactions on online platforms.
\end{abstract}





\begin{teaserfigure}
  \includegraphics[width=\textwidth]{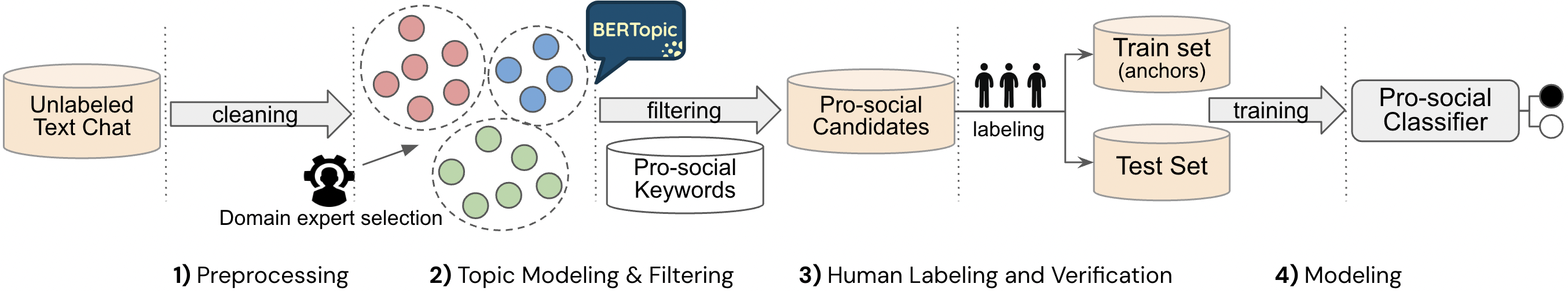}
  \caption{Text Chat Data Processing Pipeline. The  pipeline preprocesses data, uses topic modeling to identify prosocial candidates, selects candidates for labeling, and supports the development of a novel data-efficient classifier.}
  \Description{Processing steps}
  \label{fig:main_framework}
\end{teaserfigure}

\maketitle

\section{Introduction}
Analyzing the vast amounts of player communication generated on online gaming platforms is crucial for fostering positive experiences and promoting constructive interactions \citep{tomkinson2022thank}. In gaming environments, players communicate through in-game chat using \rev{a} domain-specific vocabulary. Detecting prosocial content holds the key to identifying, rewarding, and encouraging the kinds of behaviors that create a welcoming and supportive atmosphere for all players \citep{wijkstra2023help}. 
\rev{\paragraph{\textbf{Motivation}} The ability to automate the detection of prosocial behaviors also opens up possibilities for transforming game reward systems and shaping player experiences, directly incentivizing positive behaviors and supporting a healthier community atmosphere.}
\rev{Through our methodology, it becomes feasible to design reward mechanisms that directly incentivize prosocial behaviors. These rewards could include in-game currency, experience points, special cosmetic items, or other player-relevant benefits, thereby creating a positive feedback loop. Additionally, prosocial behavior detection can enable targeted interventions, identifying players with lower levels of positive interactions and suggesting tailored in-game prompts or other interventions to encourage them towards more prosocial behaviors. 
Finally, our method can support A/B testing, where developers can evaluate different prosocial reward schemes and intervention strategies, ensuring these systems have a beneficial impact on the community.}

Discovering, defining, and labeling new categories of player behavior in text chat data, such as prosocial communication, particularly those without well-established standards, pose\rev{s} a significant challenge. In the absence of pre-existing definitions or labeled examples, \rev{fine-tuning} reliable classification models becomes difficult. This problem is even more pronounced when analyzing real-world text data from gaming platforms, where the volume is vast, the underlying intent is often unclear, and domain-expert input is essential to establish consistent labeling criteria.

\paragraph{\textbf{Prior Work}} Past research and industry efforts on game chat communication have largely centered on detecting hate speech and toxicity \citep{lees2022new} – an important step for moderation and punitive intervention, such as the ones used in moderation systems \citep{makhijani2021quest}. However, in gaming domains, such a singular focus has recently been criticized for its limitations in fostering a genuinely positive and inclusive environment \citep{wijkstra2023help}. On top of that, such limited work in the area poses a technical challenge due to the lack of annotated datasets and pre-trained models for prosocial communication detection from text chat. 

\paragraph{\textbf{Our Approach}} To address these challenges, we focus on detecting prosocial behaviors in-game chat. We propose a pipeline that combines the unsupervised discovery of behavioral patterns and expert-in-the-loop input to formalize definitions and embedding-based representations for robust prosocial behavior classification.
Our subsequent classification approach introduces a novel \emph{Self-Anchored Attention Model} (SAAM) which achieves \rev{7.9\%} higher Area Under the ROC Curve (AUC)  compared to the best \rev{benchmark}. The unique aspect of our techniques lies in the efficient reuse of a small number of labeled training examples (\emph{anchors}), even in low-resource settings. Specifically, \emph{SAAM} uses the entire \rev{anchor} set as a source of features for each labeled example, \rev{and then reuses it once again} for supervised \rev{fine-tuning}. On top of that, the use of attention enables the model to most effectively combine the information from each labeled anchor example in terms of the relation it has with each \rev{anchor} example at the level of raw features extracted from game-domain adapted representations (sentence embeddings) \rev{further fine-tuned via contrastive learning}.


\paragraph{\textbf{Contributions}}
\begin{itemize} [leftmargin=*, itemsep=0.0mm, topsep=3pt]
    \item We propose \rev{prosocial player chat detection framework and a} novel Self-Anchored \rev{Attention Model (SAAM) which is a} representation-based classification method that efficiently utilizes limited labeled data.

    \item \rev{We leverage our framework and SAAM to identify prosocial player behavior dimension in real-world dataset from Call of Duty®: Modern Warfare® II, showcasing its application.}
    
    \item We characterize categories of prosocial behaviors in the gaming context such as \emph{``community building''} and \emph{``resource sharing''} by combining unsupervised discovery and collaboration with domain experts.
\end{itemize}

\section{Background and Related Work}
\subsection{Prosocial Behavior}

Multiple \rev{definitions and} types of prosocial behaviors \rev{have been} proposed in the literature. \citet{doi:10.1177/0146167209333045} define prosocial behaviors \rev{as} \textit{``those intended to help others''}. \citet{LitvackMiller1997TheSO} define it as the activity of \textit{``helping and defending''}. At the same time, \citet{FINDLAY2006347} propose a more altruism-aligned view, defining prosocial behavior as assisting or acting in someone else’s interest. Similarly,  \citet{articleeis} define prosocial behavior as voluntary, intentional behavior that results in benefits for another. \citet{articlebau} include volunteering, donating, and comforting in the prosocial category. Overall, prior work recognizes the following types of behaviors as prosocial: 
\begin{itemize} 
    \item helping and defending
    \item assisting or acting in someone else's interest
    \item volunteering, donating, and comforting
    \item cooperation and sharing
    \item expressing empathy and emotional awareness. 
\end{itemize}

Several theories attempt to explain why people can behave \textit{``prosocially.''} One theory suggests an evolutionary perspective, suggesting that prosocial behavior is a result of gene selection \citep{HAMILTON19641}, i.e., genes push us to help close relatives because it helps spread our genes. Another popular theory is reciprocity, i.e., people help others because they expect these others to help them back in the future \citep{Trivers1971-TRITEO-4}. Theories of altruism and empathy highlight the role of internal drivers in prosociality, suggesting that individuals 
may act in the interest of others out of genuine concern or a shared sense of humanity \citep{emphathy}.
However, altruism can be paradoxical as selfishness is the expected outcome when people are maximizing their utilities \citep{margolis1984selfishness}. Some literature points out that altruistic outcomes might also occur when players are selfish, such as the prisoner's dilemma, implying cooperation can happen even when players are focused only on their own interests \citep{Kuhn2008-KUHPD}. 
In summary, several empirical works in the gaming space and outside of it as well as several theoretical works provide a strong underpinning for the presence of prosocial behavior in social environments such as multiplayer games. Empirical works further provide several examples of the types of prosocial behaviors to look for in player behavior and communication.

\subsection{Toxic Content Detection}
While there is theoretical support and some evidence for the presence of prosocial behavior in a game settings,  the task of detecting prosocial behavior is underexplored compared to detecting toxic behavior, which is much more studied in the text chat context. 

Toxicity, as an opposite to pro-sociality, poses a significant challenge to fostering prosocial behavior in online gaming communities \citep{beres2021don}. Toxic behaviors, such as verbal harassment and discrimination, can significantly undermine player experiences and create hostile environments \citep{HILVERTBRUCE2020303}. To address this issue, research increasingly focuses on proactive and reactive interventions aimed at reducing toxicity. These approaches range from proactive content moderation systems that use human effort and ML algorithms \citep{makhijani2021quest} to player-centered interventions that nudge individuals towards positive interactions \citep{zora101582}. Detecting toxicity and investigating the effectiveness of various countermeasures is crucial in designing in-game reward systems that not only encourage prosociality but actively discourage negative behaviors \citep{wijkstra2023help}.

There are many existing efforts that can help detect toxic content on online platforms. For example, \emph{VaderSentiment} \citep{Hutto_Gilbert_2014} proposes a rule-based model for sentiment analysis that can help detect negative sentiment; Leong et al. \cite{leong-etal-2023-self} uses a fine-tuned \emph{Deberta-v3} model originally proposed by  \cite{DBLP:journals/corr/abs-2111-09543} to score toxic content to de-toxify pretrained language models; ToxicBert \citep{Detoxify} is another transformer-based model trained to detect various types of toxicity. Given the prevalence of off-the-shelf techniques and datasets for detecting toxicity, we leverage these in our work. Treating prosocial behavior as the direct opposite of toxicity is an obvious simplification \citep{gentile2009effects}. Prior work suggests that players can be both toxic and prosocial at the same time \citep{penner2005prosocial}. Still, toxicity models and datasets \rev{may} offer a useful prior in the absence of established equivalents for detecting prosocial behaviors.

\subsection{Unsupervised Discovery \rev{in Text}}
In the absence of readily available labeled datasets for prosocial behavior identification, we investigate unsupervised and sample-efficient supervised learning strategies.

Hundreds of thousands of players can be present in \titleFullName{} at any given time and in a 30-day period the number of unique active players can frequently reach several million \citep{kocielnik2024challenges}. Given such numbers, even if a small percentage of players use chat messaging, this can still produce vast amounts of unlabeled raw text chat. Hence, effective unsupervised methods for first organizing and exploring the vast space of such chat are needed. Additionally, the types and prevalence of prosocial behaviors are not entirely known. There are several existing methods for unsupervised clustering of text, which can help address some of these challenges. Aside from general-purpose non-text-specific clustering algorithms such as K-Means, a popular method for analysis of unlabeled text relies on topic modeling. Traditional methods like Latent Dirichlet Allocation (LDA) \citep{10.5555/944919.944937}, negative matrix factorization (NMF) \citep{10.1145/1102351.1102451} rely on discrete word frequencies and their co-occurrence (i.e., bag-of-words approach). \rev{More} recent neural-based approaches\rev{, such as BERTopic \citep{grootendorst2022bertopic}}, leverage sentence embeddings, extracted from pre-trained language models. 

\subsection{\rev{Low-Resource Setting Text Classification}}
Privacy policies on many gaming platforms often restrict the long-term storage of player-identifiable information. This well-intentioned practice safeguards user privacy but presents a challenge for prosocial behavior discovery and classification algorithm development. The expense of repeatedly relabeling large amounts of data necessitates a focus on sample-efficient supervised learning methods when working with limited manually labeled datasets.
 
There are several types of sample-efficient methods that offer promising solutions. One approach involves parameter-efficient transfer learning. The idea is to fine-tune pre-trained language models while minimizing the number of trainable parameters, like adapter \rev{tuning} \citep{houlsby2019parameterefficient}. Others use pre-trained embeddings as features to build classification models such as \emph{simCSE} \citep{gao2021simcse}, \emph{AdaSent} \citep{huang2023adasent}, and \emph{SetFit} \citep{tunstall2022efficient}. These approaches leverage sentence transformers (sBERT) to convert player communication (e.g., text chat) into meaningful embedding representations \citep{DBLP:journals/corr/abs-1908-10084}.

\section{Methods}

\subsection{Dataset Description}
We take a subset from the text chat dataset of \titleFullName{} from 2023-12-11 to 2023-12-17. The data contains sampled text chat in the public channel\rev{, which} serves as one of the real-time communication tools for players to interact with each other during gameplay ~\citep{kocielnik2024challenges}. This feature is commonly used across various multiplayer modes, allowing players to send messages to teammates or, in some settings, to all participants in a match. 

\subsection{\rev{Data Cleaning \&} Preprocessing}
\rev{Figure~\ref{fig:main_framework} provides an overview of our data processing pipeline starting with data preprocessing and cleaning (Fig. \ref{fig:main_framework}.A}) The preprocessing \rev{step} aims to enhance the quality of the dataset by eliminating text chat instances that \rev{can be trivially removed as} not \rev{containing} pro-social behavior. This process involves removing spam messages, automatically flagged toxic content, and players with very little text in the chat channel.  The steps undertaken are detailed below: 

\paragraph{\textbf{Spam Detection and Elimination:}} For each text chat entry, we employ regular expressions to detect the presence of email addresses, phone numbers, and URLs. Additionally, we calculate the length of characters for each entry. Grouping the data by player and match, we assess the sum of the length of text chat per person-match entry.  We also evaluate the variance in character length. Player-match entries are excluded if their chat includes any emails, phone numbers, or URLs. Also, we exclude ones with no standard deviation in length, since that could indicate the messages are all exactly the same.

\paragraph{\textbf{Toxic Content Filtering:}} \titleFullName{} utilizes an automatic keyword-based sanitization method, where recognized toxic words are replaced with asterisks (e.g., "****"). We use regular expressions to identify all instances of sanitized text chats and exclude them from the dataset. This is to help us focus on potential pro-social candidates. 

\paragraph{\textbf{Construction of High-Quality Player Chat Histories:}} To focus on players actively utilizing the text channel, we only consider those whose total text length exceeds 10 words across the match. The goal of this step is to make sure we have enough context for the player's communication. Individual player messages are very short with a median of 2 words in length. On top of that 54\% of the per-match communication for an individual player has a length of fewer than 5 words as shown in table~\ref{tab:hist_word}. We then compile the chat history for each player-match combination. This compiled chat history serves as the primary unit for subsequent text chat classification.

\begin{table}[!htp]
\large
\centering
\caption{Word Count Frequency of Game Chat}

\begin{tabular}{lr}
\hline
  \textbf{Word Length} & \textbf{Proportion (\%)} \\ 
  \hline
  < 5 words & 53.92 \\ 
  5-10 words & 21.17 \\ 
  10-20 words & 13.57 \\ 
  20-50 words & 7.86 \\ 
  50-100 words & 2.28 \\ 
  100-500 words & 1.17 \\ 
  500-1k words & 0.03 \\ 
  1k - 10k+ words & 0.01 \\ 
  \hline
\end{tabular}

    \label{tab:hist_word}
\end{table}

\subsection{Topic Modeling and Filtering}
 \rev{In the next step, we} perform topic modeling on the pre-processed text chat \rev{(Fig. \ref{fig:main_framework}.B)} using BertTopic \citep{grootendorst2022bertopic}, a state-of-the-art topic modeling technique that leverages \rev{sentence} embeddings to discover latent topics within a collection of documents. Following the modeling, we transform the text into embeddings with pre-trained \rev{Sentence-}BERT, and then we perform clustering. \rev{For each topic cluster}, we use TF-IDF to extract the most frequent words that represent \rev{the given topic}. 

Based on the results, we select topics best capturing pro-social behaviors in consultation with domain experts from \rev{Activison-Blizzard-King} (\company). Leveraging their extensive knowledge and understanding of the gaming community's dynamics, we aim to pinpoint themes that reflect positive social interactions. According to the insights provided by these experts, we anticipated the identification of the following five categories of pro-social behaviors:
 
\begin{itemize}
\item \textbf{Sportsmanship} - positive, respectful, trustworthy (e.g. text related to completing objectives/challenges, assist, buybacks, revives, no friendly kills)
\item \textbf{Sharing} (e.g. text related to sharing behavior - loot, resources, rides) 
\item \textbf{Good communicator} - helpful, observant, friendly (e.g. text related to pinging for nearby enemies, helping novices, calling out valuable loot - via pings/voice/text, encouraging teammates to communicate with one another)
\item \textbf{Reciprocity} (e.g., returning a revive to a teammate who revived you)
\item \textbf{Community builder} (e.g. text related to intent to play again with the same party, respawning nearby teammates, landing near teammates) 
 \end{itemize}

\begin{table}[b!]
\centering
\caption{Counts of players whose communication includes any of the pre-filtering words under each prosocial category; Note: soft match is based on the use of similar words obtained through cosine similarly from word embeddings.}

\begin{tabular}{lcccc}
\hline
\textbf{Prosocial Category} & \multicolumn{2}{c}{\textbf{Exact Match}} & \multicolumn{2}{c}{\textbf{Soft Match}} \\
 & \# players & \% total & \# players & \% total \\
\hline
Sportsmanship & 1041 & 50.4\% & 1219 & 59.1\% \\
Sharing & 403 & 19.5\% & 1079 & 52.3\% \\
Good communicator & 353 & 17.1\% & 462 & 22.4\% \\
Reciprocity & 599 & 29.0\% & 1243 & 60.2\% \\
Community Builder & 1068 & 51.7\% & 1203 & 58.3\% \\
\hline
\end{tabular}
\label{tab:player_attributes}
\end{table}
Table~\ref{tab:player_attributes} summarizes the frequency of each category among all candidate text with keyword matching, both exact matching and similar words matching (soft matching). 

The results of the topic modeling, including cluster word summaries, are presented in the Appendix (Figure~\ref{fig:topics}). Some of these categories naturally aligned with unsupervised discovery via topic modeling. We focus on the \emph{Community Builder} category due to interest from the domain experts. The community builder had the strongest presence in the discovered topics and was of most interest to the domain experts due to its positive impact related to inviting and engaging other players, which corresponds to the cluster
"0\_invite\_join\_play" in Figure~\ref{fig:topics} in Appendix \ref{apx:prosocial_keywords}.

We then select keywords based on both the topic modeling results and domain expert knowledge. A complete list of frequent keywords extracted from a cluster best aligned with the \emph{Community Builder} category \rev{are presented in Table \ref{tab:keywords}.} We report all other discovered categories and keyword lists in the Appendix \ref{apx:prosocial_keywords}.

\begin{table}
\caption{\rev{Top keywords of the discovered prosocial cluster around ``community builder''. These keywords are used for filtering content in the subsequent stages to increase the proportion of prosocial content.}}
\label{tab:keywords}. 

\begin{tabular}{llllll}
\hline
again & backup & drop & follow & guide & stay \\
team & party & play & spot & help  & invite  \\ 
join & land & near & stick & together & regroup \\ 
respawn & support & & & & \\

\hline
\end{tabular}
\end{table}

\subsection{\rev{Human Data Labeling \& Verification}}
\label{sec:human_labeling}

\rev{The third stage of our framework involves human labeling of the limited set of chat communication (Fig. \ref{fig:main_framework}-C).} The prosocial keyword filtering narrows down the candidates serving as a weak classifier.  There are multiple definitions proposed by previous literature, like \textit{helping and defending} \citep{LitvackMiller1997TheSO}, \textit{voluntary, intentional behavior that results in benefits for others} \citep{articleeis}, and \textit{volunteering, donating and comforting} \citep{articlebau}. Following previous works, we label the extracted examples according to the following definitions:

\begin{itemize}[leftmargin=*, itemsep=0.0mm, topsep=5pt]
\item \textbf{Prosocial} - \textit{``An interaction or behavior that is intended to result in a benefit to another player. This includes, but is not limited, to behaviors such as helping and defending, assisting or acting in someone else’s interest, volunteering, donating, and comforting. As well as cooperation and sharing, expressing empathy and emotional awareness.''}

\item\textbf{Unclear} - \textit{``Either not enough context to judge (i.e., can be interpreted either way), of a mix of prosocial and toxic (e.g., sarcasm).''}

\item \textbf{Not-prosocial} - \textit{``Explicitly / Implicitly toxic with an intent of offending others.''}
\end{itemize}
        
\paragraph{\textbf{Procedure}}
To implement these guidelines, we employ a labeling procedure where each data entry receives two independent labels from two \rev{human} annotators. We then assess the initial labeling agreement using 
Cohen's Kappa (0.755), which represents substantial agreement \citep{mchugh2012interrater}. Still, these agreement scores highlight a key challenge in prosocial detection: the inherent subjectivity of what different individuals consider prosocial behavior. To address this, data points with discrepancies are re-labeled with discussion between the annotators until a consensus is reached. 

\paragraph{\textbf{\rev{Final Labeled Dataset}}} The final dataset \rev{contains} 960 instances, which primarily consist of English\rev{-only} language (93.6\%). \rev{The remaining instances are a mix of English and other languages including Spanish (\~1.0\%), Persian (\~0.7\%), Russian (\~0.6\%) and other languages (see Table \ref{tab-apx:languages} in Appendix \ref{apx:non_english_counts})}. \rev{As shown in Table \ref{tab:labeled_data_stats}}, 53.0\% of \rev{labeled} instances are categorized as ``prosocial'', followed by ``not-prosocial'' (28.5\%).  A smaller portion is labeled as ``unclear'' (18.4\%). \rev{The length of text chat instances ranges from 11 to 39 words, with a mean of 30.6 and median of 32. The data is fairly equally distributed in terms of length across class labels.} We \rev{further} provide an example \rev{of the chat contents} for each label in Table \ref{tab:example_chat}. 

\begin{table}[t!]
\caption{\rev{Summary statistics for instance counts and text lengths across categories in the labeled dataset.}}
\centering
\begin{tabular}{lcccc}
\hline
 & \textbf{All Data} & \textbf{Prosocial} & \textbf{Unclear} & \textbf{Non-prosocial} \\
\hline
\textbf{Count} & 960 & 509 & 177 & 274 \\
\textbf{Percent} & (100\%) & (53.0\%) & (18.4\%) & (28.5\%) \\
\hline
\multicolumn{5}{l}{\textbf{Word Length}} \\
\hline
Min & 11 & 11 & 11 & 12 \\
Max & 39 & 39 & 39 & 39 \\
Mean & 30.6 & 30.6 & 29.9 & 30.9 \\
Median & 32 & 32 & 31 & 32 \\
\hline
\multicolumn{5}{l}{\textbf{Character Length}} \\
\hline
Min & 49 & 53 & 49 & 67 \\
Max & 747 & 310 & 215 & 747 \\
Mean & 155.5 & 154.8 & 149.9 & 160.4 \\
Median & 49 & 53 & 49 & 67 \\
\hline
\end{tabular}
\label{tab:labeled_data_stats}
\end{table}

\begin{table}[b!]
\caption{Exmaples of player text chat communicate concatenated for one player over the course of a match.}
\begin{tabularx}{\columnwidth}{p{1.2cm}  X  p{1.0cm} } 
\hline
\textbf{Keyword} & \textbf{Text} & \textbf{Label} \\ 
\hline
drop & Thank u. Yeah. Me left start fresh. One more?. Try find horse or calling card. Hvt drop a keys. He quit. Dead backpacks?where?. Bots. Careful lot bots. 4 squads. I see that. & prosocial \\ \hline
invite & bet. invite me. sniper only. sure. bet. ayoooooo. add me back. bet. bruh sniper only. bet. naw u wait for the 1v1. lmfaooo. 1v1 after this??. lol. wtf. thats an ayooooo fr fr. bruhhhhhh & unclear \\ \hline
respawn & layn down, pathetic. its a game, you respawn . 3 guys, vs me. bummys holding hands. wtf u guys are horrible. my team just hiding wtf. & not-prosocial \\ \hline
\end{tabularx}
\label{tab:example_chat}
\end{table}

\subsection{Prosocial Text Chat Classification}
\rev{The final stage of our framework involves supervised classificaiton of aggregated player text chat as ``prosocial'' or ``not-prosocial'' (Fig. \ref{fig:main_framework}-D)}.

\subsubsection{Evaluation Setup}
\rev{We frame prosocial text chat detection as a low-resource task with limited labeled data, motivated by 1) real-world constraints, where retention of large-scale annotated data is limited due to privacy requirements, and 2) the absence of existing datasets labeled specifically for prosocial behavior.}

\rev{In this setup, the training set is substantially smaller than the test set, with limited labeled examples serving as \emph{anchors} \citep{eder2020anchor}}. We split the data into a 20\% anchor set and an 80\% test set, \rev{following standard low-resource settings \citep{eder2020anchor, jiang2023low}. To estimate performance despite the small anchor set, which may lead to overfitting with full fine-tuning, we perform three seeded randomized splits and report the mean and standard deviation of performance across these splits, using consistent hyperparameters. We specifically forego extensive hyperparameter search due to high computational costs, noting also recent work showing that traditional Bayesian optimization struggles to scale effectively given the vast number of hyperparameters present in modern LLMs \citep{bamler2020augmenting}. We instead perform model averaging across folds as a robust alternative to traditional tuning methods as proposed in \cite{schmidt2023one}.
All experiments have been performed on a single A100 GPU with 85GB of memory.}

\rev{For prosocial content detection, we cast the task as a binary classification by combining \textit{``not-prosocial"} and \textit{``unclear''} into a single \textit{``not prosocial''} category. This approach balances the class labels, simplifies the task, and better aligns with the practical goal of detecting prosocial versus other content.}

\subsubsection{Metrics}
\rev{We report the mean and standard deviation of Area Under the ROC Curve (AUROC), F1 Binary (F1-bin), Precision, and Recall across three randomized splits. AUROC assesses the model's ability to distinguish between classes, with 1.0 indicating perfect classification. The F1-bin score balances precision and recall, providing a useful measure of the model's ability to correctly identify prosocial content while minimizing false classifications.
Precision measures the accuracy of prosocial predictions, while Recall indicates the proportion of actual prosocial instances correctly identified. These metrics provide a robust performance estimate, accounting for variability from the limited anchor set.}

\subsubsection{\rev{Benchmark Models}}
We compare to several baselines. As there are no pre-trained models for prosocial content detection, we leverage pre-trained models for sentiment and toxicity detection. We further fine-tune transformer architectures and \rev{Conv}LSTM models on our dataset. Finally, we compare our model to several lightweight pre-trained representation-based techniques.

\paragraph{\textbf{\rev{Frozen Models and Few-Shot Prompting Approaches}}} We compare to the existing sentiment and toxicity models. For the sentiment models, we treat the positive sentiment label as a proxy for the prosocial label. We treat the non-toxic label for the toxicity model similarly. We compare to the popular rule-base \emph{VaderSentiment} model \citep{elbagir2019twitter}. We further use the best-performing dimensions of \textit{``toxicity''} and \textit{``insult''} from pre-trained \emph{ToxicBert} \citep{Detoxify}. Finally, we use one of the latest best-performing pre-trained toxicity detection models - ``deberta-v3-large-toxicity-scorer'' \emph{(Deberta-v3-large-tox)} from \cite{leong2023self}. We further also test the few-shot (k=4 worked best) prompting of general-purpose models that can be run locally, such as \emph{Mistral-7B-v0.1} and \emph{Mistral-7B-v0.2} \citep{jiang2023mistral}. We used the class-balanced shots selected by cosine similarity using pre-trained SBERT \textit{``all-MiniLM-L12-v2''} \citep{reimers2019sentence} similarly to prior work in \cite{prabhumoye2021few} and \cite{kocielnik2023can}.

\paragraph{\textbf{Fine-tuned Models}}
We finetune \rev{two sizes of bert family models \citep{devlin2018bert}},  \emph{``bert-base-uncased''} (\emph{BERT}) \rev{and \emph{``bert-large-uncased''} (\emph{BERT-LG})} using standard HuggingFace trainer \citep{Trainer26:online} for 5 epochs \rev{with the default learning rate (LR) of $5e^{-5}$, batch size of 16, weight-decay of 0.01, and 10 warmup steps. These represent the default settings for the trainer, with only the number of epochs adjusted based on empirically observed convergence.}

We further leverage the adapter framework\rev{, due to its parameter-efficient nature suitable for small data \citep{hu2023llm, le2024impact}. Using adapter tuning, we} fine-tune the same \rev{bert-base} model (\emph{BERT + adapter}) as well s high zero-shot performing models \emph{(ToxicBert + adapter)} \rev{and \emph{(Deberta-v3 + adapter)}}. \rev{In all adapter finetuning we use LR of $5e^{-5}$, batch size of 16, warmup of 10 steps and weight decay of 0.01. These represent the default settings for the trainer.}

Finally, we also train a custom ConvLSTM architecture with linear attention \citep{firoz2023automated}, which embeds individual chat messages using a pre-trained SBERT embedding (\textit{``all-MiniLM-L12-v2''}) and combine these representations as a sequence. This was inspired by the work on toxicity detection from in-game chat modeling individual messages as a sequence of conversations \citep{weld2021conda}. \rev{Due to the custom nature of this architecture and the lack of a default HuggingFace trainer, we experimentally determined the LR of $1e^-3$ and learning epochs of 30.}

\paragraph{\textbf{Representation-based Approaches}}
We also compare our approach to several lightweight methods that leverage or adjust pre-trained representations in a data-efficient manner. Specifically, we compare to unsupervised and supervised variants of \emph{SimCSE} \citep{gao2021simcse}, as well as to \emph{SetFit} \citep{tunstall2022efficient}. \rev{For these approaches, we used a limited grid-search following \citep{gao2021simcse} with batch size $\in$ $\{$16, 32, 64$\}$ and learning rate $\in$ $\{$1e-5, 3e-5, 5e-5$\}$, with the number of training epochs set to 1.} Finally, we compare to a simple KNN, \rev{for which} k=24 worked the best \rev{over a search across k$\in$$\{$2, 8, 12, 24, 32, 48, 64$\}$} approach leveraging the mean cosine-similarity to the closest positive and negative anchors using pre-trained SBERT embeddings. 

\begin{figure*}[t!]
\includegraphics[width=1.\textwidth]{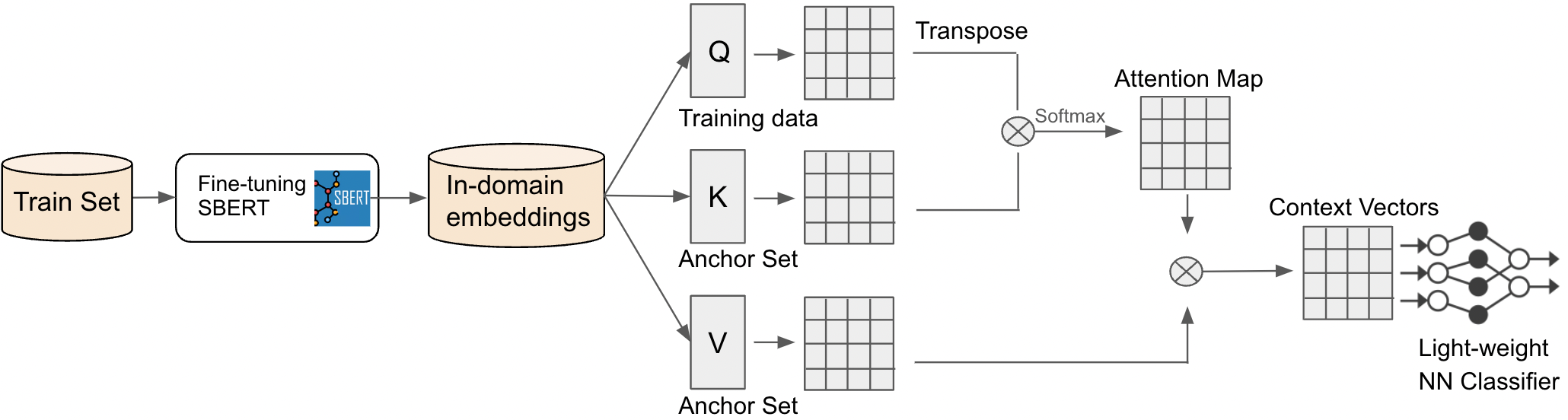}
\caption{Self-anchored Attention Model. Note that for our self-anchoring models, the anchor set is \rev{used twice, first for embedding representation fine-tuning and later for training the mult-head attention architecture}.}
\label{fig:embedding}
\end{figure*}

\subsubsection{Self-Anchored Attention Model}

\rev{As shown in Fig. \ref{fig:embedding}} our approach extends representation-based classification methods, with a key focus on efficiently utilizing the limited \emph{``anchor''} set. To maximize the information from the \rev{anchor} set, we leverage it twice. First, for representation learning through embedding fine-tuning to the player chat using contrastive loss. The second time, for \rev{classification model fine-tuning with attention. We describe these two core stages of our approach.}

\paragraph{\textbf{Step 1: Embedding \rev{Fine-tuning}}}
Game chat data presents a distinct challenge for general-purpose language models due to its specialized vocabulary (slang, abbreviations, and game-specific terminology), informal style (misspellings, grammar variations), and nuanced semantic relationships based on game-related events and actions \citep{furianto2023gaming}. Fine-tuning a sentence transformer model (SBERT) directly addresses these challenges by adapting the embedding representation to the unique linguistic characteristics of in-game conversations. 

We fine-tune the sentence transformer with the \rev{anchor} set using contrastive loss \citep{khosla2020supervised}. The idea for contrastive loss is to learn an embedding space where similar examples are placed closer together, while dissimilar examples are pushed further apart. This is achieved by minimizing the distance between positive pairs (similar examples) and maximizing the distance between negative pairs (dissimilar examples).

For positive sampling, we randomly sample 10 anchors from the \rev{anchor} set for each pro-social keyword and for each label. That is 10 prosocial anchors and 10 non-prosocial anchors for each pro-social keyword. Note that in this setting, we might oversample the \rev{anchor} set given the small number of \rev{labeled} examples. Then, for each keyword, we use all pairwise combinations of prosocial anchors and all pairwise combinations of non-prosocial anchors as positive samples.
That is, we have $2\ *$ $10 \choose 2 $ positive pairs. For negative pairs, we pair within each keyword all prosocial anchors with all non-prosocial anchors. That gives us $10 * 10$ negative pairs per keyword. 

By fine-tuning SBERT on this in-context \rev{game chat} language, the model becomes far better equipped to handle key tasks within the gaming domain.

\paragraph{\textbf{Step 2: Classification Modeling with Self-Anchoring}}
To overcome the limited number of labeled examples, we propose a novel approach to classification modeling with the idea of self-anchoring. This method takes input from the embeddings derived from a fine-tuned sentence transformer. The core of the self-anchoring approach is the use of the whole \rev{anchor} set for \rev{fine-tuning}, utilizing it in its entirety as input features. Such a technique enables the model to learn the nuanced relationship between \rev{all examples in the anchor set}  and the \rev{particular individual anchor} example to label.

We tried multiple approaches using the self-anchoring idea. Below are two approaches that give the best results compared to other off-the-shelf models or standard approaches. 

\paragraph{Self-Anchoring Attention Model: QKV \& K attention}

In order for the model to learn how to utilize the anchor set, we attempt the attention mechanism. Here, we take the whole \rev{anchor} set and build an attention layer using self-attention. After doing the transformation, we pass it through a classification head (dense layer). Fig.~\ref{fig:embedding} shows the structure of the model. It has the following components: 

\begin{enumerate}
    \item Query (Q), Key (K), Value (V) Transformations: Separate dense layers transform the input embeddings into queries, keys, and values. The query transformation is applied to the input embeddings x, while the key and value transformations are applied to the anchors.
    
\item Attention Scores: The dot product of queries and keys generates attention scores. These scores determine how much focus each element of the input sequence should receive. A softmax function is then applied to the attention scores to ensure they are normalized (i.e., sum up to 1 across the keys for each query).
$$
\text{AttentionScores} = \text{softmax}\left(\frac{QK^T}{\sqrt{d_k}}\right)$$

\item Context Vector: The context vector is a weighted sum of the values, where the weights are given by the attention scores. This vector is then used for further processing or as an output for classification tasks.

$$\text{ContextVector} = \text{AttentionScores} \cdot V$$

\end{enumerate} 
The attention mechanism's ability to weigh different parts of the anchor embedding differently is crucial in a setting where the difference between prosocial and non-prosocial content can be subtle and context-dependent. It enables the model to focus on linking \rev{anchor} data with both prosocial and non-prosocial anchors, in this case, the \rev{anchor} set itself, improving its accuracy with relatively few  examples by utilizing the \rev{anchor} data extensively. 

When testing the approach, we encountered overfitting when \rev{fine-tuning on} more batches. To reduce overfitting, we introduce a simplified version of attention, which only learns the weighting matrix of K and ignores the weighting matrix for Q and V. The simplified version of the model preserves a similar AUC score, while greatly reducing overfitting with more training epochs.

\paragraph{Self-Anchoring Attention Model: Cosine Similarity}

Building upon the self-anchoring concept, this approach leverages cosine similarity to capture the inherent relationships between \rev{features of all anchor} examples \rev{and each individual example to label}. We calculate the cosine similarity between each \rev{anchor} embedding and all other embeddings in the \rev{anchor} set. This essentially captures how ``similar'' each example is to all the anchors.  By incorporating these similarity scores as additional features,  we enable the model to exploit the underlying structure and relationships within the \rev{anchor set}. This strategy aligns with the self-anchoring \rev{approach} of utilizing the entire \rev{anchor} set to inform the classification task.

A fully-connected neural network (FC-NN) is then employed to process these self-anchoring similarity scores. The FC-NN allows the model to learn complex, non-linear relationships between the similarity features and the target label (prosocial or non-prosocial) with a non-linear activation function. 
Finally, a classification head predicts the class for each input sample. 

\begin{table*}[t!]
\large
\centering
\caption{AUC Scores Comparison Across Different Methods. Note: we report the mean AUC score over 3 randomized splits of train/test sets. We report \% gain/loss over the best existing technique. * denotes variations of our novel \emph{SAAM} model. }

\begin{tabular}{lccccc}
\hline
\textbf{Method} & \textbf{AUROC (SD)} & \textbf{Gain/Loss} & \textbf{F1-bin} & \textbf{Precision} & \textbf{Recall}\\
\hline
VaderSentiment & $.614_{\pm.007}$ & \dar{20.8\%} & \rev{$.663_{\pm.005}$} & \rev{$.608_{\pm.010}$} & \rev{$.730_{\pm.003}$} \\

Deberta-v3-large-tox & $.726_{\pm.002}$ & \dar{6.3\%} & \rev{$.741_{\pm.004}$} & \rev{$.737_{\pm.010}$} & \rev{$.747_{\pm.005}$}  \\

ToxicBert (toxicity) & $.666_{\pm.003}$ & \dar{14.1\%} & \rev{$.686_{\pm.011}$} & \rev{$.671_{\pm.013}$} & \rev{$.700_{\pm.008}$}  \\

ToxicBert (insult) & $.603_{\pm.005}$ & \dar{22.2\%} & \rev{$.728_{\pm.009}$} & \rev{$.578_{\pm.010}$} & \rev{$.982_{\pm.002}$}  \\

Mistral-7B-v0.1 & $.624_{\pm.024}$ & \dar{19.5\%} & \rev{$.700_{\pm.017}$} & \rev{$.608_{\pm.029}$} & \rev{$.826_{\pm.035}$} \\

Mistral-7B-v0.2 & $.652_{\pm.003}$ & \dar{15.9\%} & \rev{$.617_{\pm.007}$} & \rev{$.710_{\pm.030}$} & \rev{$.547_{\pm.029}$} \\

\midrule
BERT-finetuning & $.747_{\pm.028}$ & \dar{3.6\%} & \rev{$.753_{\pm.016}$} & \rev{$.727_{\pm.055}$} & \rev{$.788_{\pm.076}$} \\

\rev{BERT-LG-finetuning} & \rev{$.740_{\pm.015}$} & \dar{4.6\%} & \rev{$.761_{\pm.019}$} & \rev{$.735_{\pm.023}$} & \rev{$.788_{\pm.015}$} \\

BERT+adapter & $.714_{\pm.030}$ & \dar{7.9\%} & \rev{$.754_{\pm.008}$} & \rev{$.703_{\pm.046}$} & \rev{$.821_{\pm.071}$}  \\

ToxicBert+adapter & $.761_{\pm.005}$ & \dar{1.8\%} & \rev{$.791_{\pm.012}$} & \rev{$.744_{\pm.022}$} & \rev{$.845_{\pm.015}$} \\

\rev{Deberta-v3+adapter} & \rev{$.775_{\pm.014}$} & - & \rev{$.793_{\pm.008}$} & \rev{$.767_{\pm.047}$} & \rev{$.825_{\pm.050}$} \\

ConvLSTM (SBERT) & $.755_{\pm.007}$ & \dar{2.6\%} & \rev{$.785_{\pm.003}$} & \rev{$.747_{\pm.015}$} & \rev{$.828_{\pm.023}$} \\
\hline

SetFit (SBERT) & $.769_{\pm.025}$ & \dar{0.7\%} & \rev{$.785_{\pm.021}$} & \rev{$.850_{\pm.053}$} & \rev{$.732_{\pm.0.48}$} \\

SimCSE (unsupervised) & $.707_{\pm.016}$ &\dar{8.8\%} & \rev{$.728_{\pm.002}$} & \rev{$.730_{\pm.010}$} & \rev{$.725_{\pm.014}$}  \\

SimCSE (supervised) & $.733_{\pm.013}$ &\dar{5.5\%} & \rev{$.747_{\pm.008}$} & \rev{$.740_{\pm.012}$} & \rev{$.755_{\pm.015}$}  \\

KNN (k=24, SBERT) & $.751_{\pm.011}$ & \dar{3.1\%} & \rev{$.753_{\pm.016}$} & \rev{$.754_{\pm.019}$} & \rev{$.752_{\pm.015}$}  \\

\midrule

$^{*}$SAAM (Cos-Sim) & $.810_{\pm.008}$ & \uab{4.6\%} & \rev{$.800_{\pm.003}$} & \rev{$.788_{\pm.004}$} & \rev{$.813_{\pm.008}$}  \\

$^{*}$SAAM (K attention) & $.836_{\pm.012}$ & \uab{7.9\%} & \rev{$.811_{\pm.003}$} & \rev{$.807_{\pm.008}$} & \rev{$.816_{\pm.013}$} \\

$^{*}$SAAM (QKV attention) & $.830_{\pm.005}$ & \uab{7.1\%}  & \rev{$.799_{\pm.014}$} & \rev{$.758_{\pm.023}$} & \rev{$.845_{\pm.006}$}  \\
\hline
\end{tabular}
\label{tab:clf-results}
\end{table*}

\section{Results}
In Table  \ref{tab:clf-results} we report the results of binary classification of prosocial category - \emph{``Community Builder''} based on different pre-trained, fine-tuned, and sample-efficient methods.

Among the tested methods, The \emph{SAAM} (K attention) model achieved the highest AUC score, with a mean of $0.836\rev{\pm0.012}$ over three folds, closely followed by the \emph{SAAM} (QKV attention) model at $0.830\rev{\pm0.005}$. Using cosine similarity to all the anchors as features with an NN approach also yielded strong results, scoring an AUC of $0.810\rev{\pm0.008}$. These approaches outperformed other methods, including various pre-trained models like \emph{Deberta-v3-large-tox}, \emph{ToxicBert}, with \emph{Deberta-v3} reaching an AUC of $0.726\pm0.002$, and \emph{ToxicBert} an AUC of $0.666\pm0.003$.

Those approaches are then outperformed by \emph{SetFit} which fine-tunes the sentence transformer embeddings to the target domain and achieves the highest mean AUC of $0.769\rev{\pm0.025}$ \rev{and \emph{Deberta-v3+adapter}, which achieves an AUC of $0.775\pm0.014$.} These are the highest scores among all the competitive approaches. \emph{SetFit} also surpasses Bert-finetuning and its adapter version, which achieved AUC scores of $0.747\rev{\pm0.028}$ and $0.714\rev{\pm0.030}$ respectively. It also slightly surpasses \emph{ToxicBert} fine-tuned using adapter tuning which achieves an AUC of $0.761\rev{\pm0.005}$. We also tried \emph{SimCSE} for fine-tuning the embedding model, with the unsupervised version achieving an AUC of $0.707\rev{\pm0.016}$ and the supervised version achieving $0.733\rev{\pm0.013}$. The high performance of \emph{SetFit} among our baselines makes sense, given that this approach was specifically developed to fine-tune sentence embeddings efficiently \rev{with few labeled examples}. This comparison to several competitive approaches highlights the effectiveness of our models, particularly the QKV attention and K attention mechanisms, in accurately detecting prosocial text based on the minimal labeled training data available. 

The gain in our approach comes from its self-referential learning mechanism. By treating the entire \rev{anchor} set as both the source of anchor texts for training and input features, the model gains exposure to a broad spectrum of linguistic expressions in various contexts, both prosocial and not-prosocial. This comprehensive familiarity allows the model to internalize the \rev{complex} relationships between different message types and the characteristics that define pro-social content. Essentially, each example serves as a reference point, guiding the model's learning process by illustrating examples of language use that either aligns with or diverges from pro-social norms.
Furthermore, the self-anchoring technique enriches the model's input space, offering a denser, more informative feature set from which to learn. In short, this method not only overcomes the challenges posed by limited labeled data but also ensures that the model effectively uses the information in the \rev{all the} labeled examples.

\subsection{Ablation Studies}
To isolate the impact of each step in our pipeline, we performed ablation studies \rev{as shown in Table~\ref{tab:ablation}}. This involved removing individual components and observing the effect on the overall performance.

Removing the SBERT embeddings fine-tuning stage resulted in a decrease in AUC of 2.40\%, indicating that 
\rev{gaming languague is somewhat different than the language on which embeddings were pretrained, hence specilizing the representations to this language} improves its effectiveness for our task. A more significant drop in AUC (9.12\%) was observed when we removed the attention mechanism entirely. This highlights the critical role of the attention mechanism in our approach, as it allows the model to focus on the most relevant parts of the embeddings for classification.

Reducing the \rev{anchor} set size from \rev{20\%} to 10\% results in a decrease in AUC of 16.9\%, while reducing the \rev{anchor} set size from \rev{20\%} to 15\% results in an AUC decrease of 4.8\%. This shows the marginal gain we have from using \rev{a larger anchor set}.

\begin{table*}[th!]
\large
\centering
\caption{Ablation Study on individual parts of \rev{our Self-Anchored Attention Model}.}
\begin{tabular}{>{\raggedright\arraybackslash}lccccc} 
\hline
\textbf{Method} & \textbf{AUROC (SD)} & \textbf{Change} & \textbf{F1-bin} & \textbf{Precision} & \textbf{Recall} \\
\hline
SAAM w/o attention & $.757_{\pm.005}$ & \dar{9.1\%} & \rev{$.765_{\pm.008}$} & \rev{$.755_{\pm.011}$} & \rev{$.744_{\pm.007}$} \\
SAAM w/o fine-tuning &  $.813_{\pm.010}$ & \dar{2.4\%} & \rev{$.796_{\pm.006}$} & \rev{$.791_{\pm.003}$} & \rev{$.800_{\pm.010}$} \\
SAAM (QKV, 10\% train) & $.690_{\pm.036}$ &  \dar{16.9\%} & \rev{$.735_{\pm.009}$} & \rev{$.725_{\pm.008}$} & \rev{$.745_{\pm.011}$} \\
SAAM (QKV, 15\% train) & $.790_{\pm.051}$ & \dar{4.8\%} & \rev{$.751_{\pm.003}$} & \rev{$.745_{\pm.002}$} & \rev{$.757_{\pm.005}$} \\
SAAM (QKV, 20\% train) & $.830_{\pm.005}$ & - & \rev{$.799_{\pm.014}$} & \rev{$.758_{\pm.023}$} & \rev{$.845_{\pm.006}$}\\
\hline
\end{tabular}
\label{tab:ablation}
\end{table*}

\subsection{Additional Analysis and Manual Inspection}

\begin{figure}[t!]
    \includegraphics[width=\columnwidth]{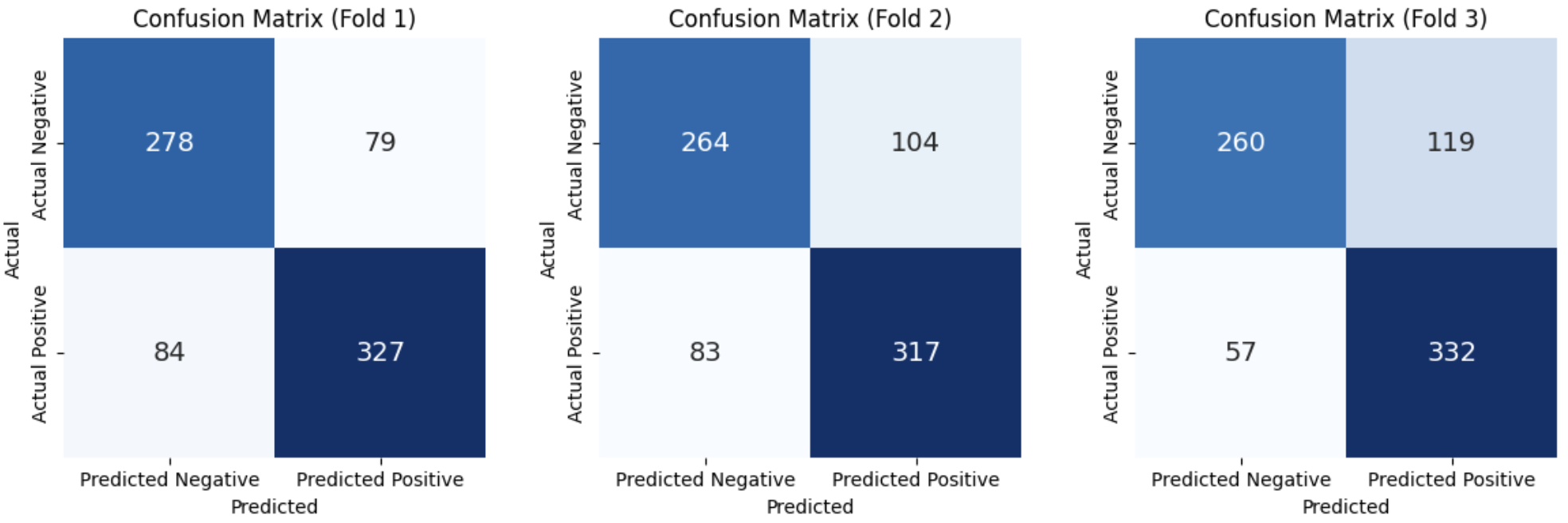}
    \caption{Confusion Matrices for the test set across the 3 data folds for the best SAAM model. The ``prosocial'' label is coded as Positive, while ``not-prosocial'' as Negative. We note fairly balanced precision and recall of the model.}
    \label{fig:cv_saam_folds}
\end{figure}

\begin{figure}[t!]
    \includegraphics[width=\columnwidth]{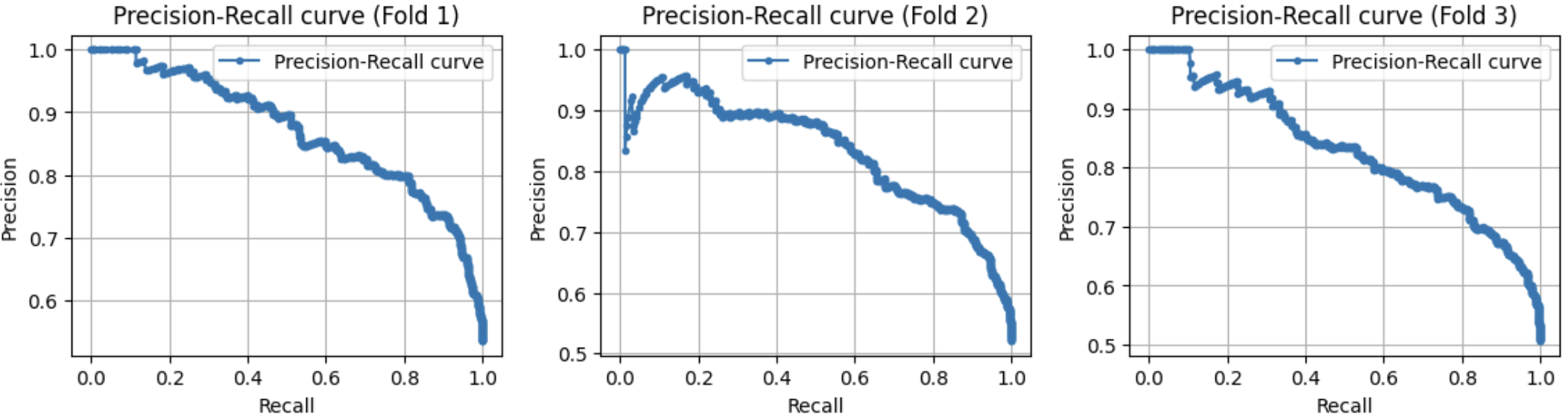}
    \caption{Precision-recall curves for the test set across the 3 data folds for the best SAAM model.}
    \label{fig:pr_saam_folds}
\end{figure}

\rev{We further inspected the test set examples labeled by our SAAM model. The confusion matrices across 3 data splits are presented in Figure \ref{fig:cv_saam_folds}. We see that the model avoids false positive and false negative equally well, with only a slight preference towards avoidance of False Negatives, which in practical application can still be balanced by adjusting the classification threshold. To explore the impact of such different classification thresholds, in Figure \ref{fig:pr_saam_folds} we plot the precision-recall curves on a test set for each data split. Here, we also see a fairly stable trade-off, with Precision \~90\% achievable under recall of 30\%-40\%. Similarly, a precision of \~80\% is achievable under recall of 60\%-80\% across data splits.}

\rev{Furthermore, we manually inspected the classification discrepancies between human annotators and the SAAM model. We identified several distinct categories of mislabeling. Table \ref{tab:mislabeling} presents these categories along with representative examples extracted from the player chat dataset. The categories include `Ambiguity in Language Use', where misinterpretations arise from positive words or friendly slang taken out of context; `Use of Gaming Jargon and Slang', where specialized gaming terms are likely misunderstood by the model; `Short and Fragmented Messages', which may lead to errors from insufficient context; `Emotional Expressiveness Misinterpreted', where emotional expressions are most likely misconstrued; and `Sarcasm or Irony', which the model may fail to detect well because player sentiment might change across the entirety of a match.}

\rev{Given that our manual inspection revealed that prominent misclassification reasons were due to various forms of ambiguity and mixed sentiment, we further examined the distribution of instances labeled as `unclear', based on human labeling, among our confusion matrix regions. We leveraged the fact, as noted in \Sref{sec:human_labeling}, that the original human labeling included an explicit category of `unclear', which was later labeled as `not-prosocial' in the subsequent binary classification formulation. We found that between 57\% and 71\%, depending on the dataset split, of the false positives were originally annotated as `unclear' by human annotators. Given that only 11\%-16\% of the labeled test set was categorized as false positives (see Figure \ref{fig:cv_saam_folds}), and only 18\%-20\% of the test set data is labeled as 'unclear' (see Table \ref{tab:labeled_data_stats}), this suggests that the instances mislabeled by the model represent genuinely ambiguous cases.}

\begin{table}[h]
\small
\centering
\begin{tabular}{p{4cm}p{4cm}}
\hline
\textbf{Category} & \textbf{Examples} \\
\hline
\textbf{Ambiguity in Language Use} - Misinterpretations due to positive words or friendly slang taken out of context. & ``learn to move in a respawn mode.'' $\bullet$ ``this cant be real. wait. wait for us to respawn bruh.'' $\bullet$ ``gg. EZ noobs. sorry. let's play again!'' \\
\hline
\textbf{Use of Gaming Jargon and Slang} - Inaccurate interpretations stemming from specialized gaming terms used in complex contexts. & ``aim alone. go res.'' $\bullet$ ``one respawn our. shield'' $\bullet$ ``the play again button broke'' $\bullet$ ``will kill u again nooby.'' \\
\hline
\textbf{Short and Fragmented Messages} - Short, fragmented messages might lack sufficient context for accurate interpretation, which can lead to errors. & ``land on me quaq. still alive quaq?. good. stay dead. you thought.'' $\bullet$ ``omg. where are you guys. omg. okokoko. i got it. :)))))). got a uav.'' $\bullet$ ``he is good. hello. omg. amk. nice try. omg.'' \\
\hline
\textbf{Emotional Expressiveness Misinterpreted} - Emotional outbursts or playful banter, especially with mixed signals like insults or frustration mixed with positive affirmations. & ``you take longer. well done.'' $\bullet$ ``drive away. team of 4 can't do shit.'' $\bullet$ ``yh imma 1 hit wonder lool gg yall. ill play respawn a little. '' \\
\hline
\textbf{Sarcasm or Irony} - Challenges in detecting positive words used sarcastically or ironically & boy obviously. ok go.'' $\bullet$ super helpful, not.'' $\bullet$ 11doc u good?. go stick 2 shipment bro. more of ur pace.''\\
\hline
\end{tabular}
\caption{Categories of Mislabeling with Representative Examples. We separate examples for different player matches using $\bullet$ for easier readability.}
\label{tab:mislabeling}
\end{table}


\section{Discussion}
\rev{Our work contributes a pipeline and novel classification method to prosocial behavior detection in online gaming. Our classifcation approach is effective particularly in resource-constrained environments with limited labeled data. 
The broader impact of this work is significant. Firstly, it shows the existence of prosocial behavior in competitve gaming. Secondly, it offers an effective practical way to detect such behavior from real-world gaming chat data. Finally, it offers insights into categories of prosocial behaviors as well as their prevalence in game chat.}

\rev{Ultimately this work enables a shift in moderation practices from solely mitigating toxic interactions to actively promoting prosocial behaviors. Our work allows gaming platforms to encourage community-building and cooperative behavior, thereby enhancing player experience and fostering healthier online communities. Nevetheless, several aspects of our approach as well as its future use warrant further discussion.}

\subsection{Impact of Leveraging Pre-trained Language Representations}
\rev{Advanced language modeling in gaming is challenging due to sparse, fragmented chat interactions further influenced by in-game events often not captured in text \citep{weld2021conda}. These characteristics make it difficult to capture nuanced, contextual language. Pre-trained embeddings from LLMs, learned from diverse datasets, offer a solid foundation to generalize across varied contexts \citep{chen2020recall}. Yet, the distinctive language in online gaming—rich in slang, abbreviations, and typos—necessitates additional fine-tuning to align general representations with gaming-specific nuances \citep{fang2023cert}. Fine-tuning is particularly important for underrepresented language patterns, as models trained only on limited labeled game chat may miss less common language use \citep{chen2020recall}. Leveraging LLM embeddings pre-trained on broad corpora enables the model to better handle these challenges, enhancing performance on underrepresented prosocial content and potentially improving fairness \citep{reddy2022benchmarking}.}

\rev{Nonetheless, pre-trained embeddings may inherit social biases present in the original general domain training data. While biases in LLMs are well-documented \citep{gallegos2024bias}, targeted fine-tuning on gaming data using contrastive learning techniques can help mitigate these inherited biases, enhancing the model’s fairness in downstream applications \citep{jin2021transferability}. Also the biases in general domain LLMs are more thoroughly studied than those in gaming text, allowing for more transparent bias detection \citep{kocielnik2023biastestgpt} and data-efficient mitigation strategies \citep{han2024chatgpt}. While the lack of player demographic data limits our ability to assess fairness directly across different social groups, we emphasize that missing an opportunity to reward a prosocial behavior under high precision detection has arguably lower negative impact than incorrectly rewarding or missing the detection of toxic behavior.}

\subsection{Nuances of Interpreting Prosocial Behavior in Competitive Game Chat}
\rev{Identifying prosocial behavior in competitive gaming is challenging due to the context-dependent dynamics of player interactions as well as annotator-subjective interpretations. Our observations reveal that prosocial behaviors in games are often open to interpretation, with competitive settings adding further ambiguity. For example, a compliment might be sincere or sarcastic, depending on context and community norms \citep{poeller2023suspecting}. Similarly, players’ pleas for help may invite prosocial responses or be seen as nagging if repeated. Phrases like ``gg'' (good game) may express sportsmanship or sarcasm, complicating automated classification efforts. Similar contextual challenges are observed in toxicity detection, where studies such as \cite{beres2021don} and \cite{yang2023towards} highlight the cultural nuances of "trash-talk," which may be accepted within certain communities but difficult to separate from negative intent. These complexities extend to interpreting prosocial behaviors, a relatively unexplored area that our work addresses by providing a foundational approach and baseline for future research.}

\rev{Furthermore, as noted by \cite{xiao2023supporting} and \cite{chen2018using}, interpretations of qualitative data will always retain a degree of subjective judgment, leading to potential variations even among trained annotators. Part of the fundamental challenge comes from inherently limited context  \citep{deterding2021flexible}. To reduce ambiguity, we used a collaborative approach, combining unsupervised topic modelingm consultations with domain experts, and using independent human annotations for labeling. By refining coding guidelines, we achieved substantial inter-rater agreement that provides high-quality training data. Further efforts can be made on the algorithmic side in the form of better methods for uncertainty quantificaiton, which represents an active area of research in deep learning NLP \citep{farr2024llm}.}


\subsection{\rev{Integrating Prosocial Behavior Detection into} Reward \rev{Mechanisms}}
\rev{The ability to automatically detect prosocial behaviors opens new possibilities for transforming game reward systems and shaping player experiences. Reward mechanisms in gaming are well-documented as powerful tools for influencing player motivation and retention, commonly through intrinsic rewards like personal satisfaction and extrinsic rewards such as badges, levels, and special items \citep{frommel2022daily}. Studies show that immediate rewards for actions like healing or assisting teammates can effectively reinforce cooperative behaviors \citep{liu2015short}. Although our system has not yet been integrated with such reward mechanisms, it holds promising potential to inform real-time recognition systems that could encourage prosocial actions, creating a more supportive and positive gaming environment.}

\rev{Our approach is adaptable to the evolving language and expressions of prosocial behavior in games \citep{seiffert2018memes}. By using a data-efficient approach that requires few labeled examples, it efficiently updates with shifting definitions, a crucial feature for future reward systems to avoid misclassification or unintended rewards. This data-efficient design allows SAAM to adapt without extensive human intervention. As such, our prosocial detection framework offers a strong basis for future reward mechanisms aligned with community values, paving the way for more inclusive and community-focused gaming experiences.}

\subsection{Privacy Challenges around the use of Third-party LLMs}

While \rev{black-box, third-party} LLMs\rev{, such as GPT-4o \citep{achiam2023gpt} or Gemini \citep{team2023gemini}}, offer broad language capabilities, our lightweight pro-social classifier excels by addressing specific concerns within gaming environments.
\rev{Importantly, our framework respects privacy by relying on locally trained models, keeping data within secure platform domains and reducing risks of data breaches \citep{bibi2024legal}.}
\rev{As such, it also} enables people to carefully curate the \rev{labeled} dataset, ensuring player communications are handled with sensitivity. Our model's focused design increases efficiency – it analyzes in-game interactions quickly and uses fewer computational resources compared to the demands of LLMs. Most importantly, it's trained on data specifically reflecting the nuances of in-game communication, giving it a deeper understanding of pro-social behaviors within the gaming domain. Additionally, LLMs often refuse to work when dealing with toxic content, whereas our specialized model can be designed to address such issues directly.  Finally, our approach allows for clear replicability and is more transparent and interpretable than use of commercial LLMs.

\subsection{\rev{Future Integration with Active Learning and Semi-Supervised Learning}}

\rev{A promising technical extension of our framework involves integrating \emph{active learning} and \emph{semi-supervised} approaches to potentially boost model accuracy, particularly in low-resource setting.}

\rev{Active learning is a method that engages human experts iteratively to label highly informative examples, maximizing data utility. However, its effectiveness depends on a high-quality initial dataset to prevent cold-start issues \citep{bodenstedt2019active}. Our method complements active learning by constructing such a dataset. Moreover, active learning’s iterative refinement requires accurate uncertainty estimates, yet deep learning models often produce overconfident predictions, complicating sample selection \citep{laparra2021review}. Finally, active learning demands frequent retraining, which is computationally intensive for deep learning models. Further work is needed to address these challenges.}

\rev{Semi-supervised learning (SSL) offers an opportunity to expand the effective training dataset by utilizing a large amount of unlabeled game chat data alongside a smaller, high-quality labeled dataset. However, SSL has limitations, such as susceptibility to noise in unlabeled data, which may impair performance if irrelevant information is included \citep{reddy2018semi}. Moreover, SSL’s success depends on labeled data quality and representativeness, as an unrepresentative labeled set could hinder model generalization \citep{oriola2020improved}. Our framework supports building robust labeled datasets, complementing SSL approaches offering an avenue for future work.}

\subsection{Limitations}
\rev{As we discussed,} pretrained language embeddings can introduce social biases, potentially affecting the fairness of the prosocial detection system\rev{, while fine-tuning for the domain can reduce the impact of such biases, it may also introduce new, domain specific ones. Future work is needed to investigate this aspect more}. Additionally, prosocial behavior can contain inherent ambiguities, making interpretation challenging even for humans, \rev{while, as disussed, we took steps to reduce these issues, we acknowledge more work is needed here}. When deployed in reward systems, the classifier carries the risk of false positives and false negatives, potentially leading to unintended consequences for players. It's crucial to \rev{account for} these limitations and discuss ethical considerations like transparency, accountability, and the potential impact on player autonomy to ensure the responsible and beneficial use of \rev{any} prosocial detection system, \rev{including ours}.

\section{Conclusion}
\rev{This study introduces a novel, self-anchored attention model (SAAM) for detecting prosocial behaviors in online game chats, specifically in Call of Duty: Modern Warfare II. Our approach addresses the challenge of prosocial detection, a new focus in competitive gaming traditionally centered on toxicity, by collaborating with domain experts to define a taxonomy of positive behaviors, such as “community building,” and demonstrating their presence in player interactions. Given limited labeled data for training prosocial detection models, SAAM uses self-anchoring, where the training set itself serves as reference points, enabling efficient learning from minimal examples and precise identification of nuanced prosocial behaviors. By shifting moderation from punitive measures to actively encouraging positive interactions, this framework supports the development of reward systems that promote constructive engagement. SAAM’s design may also extend beyond gaming, offering applications in other digital spaces where fostering prosocial interactions is critical, contributing to more inclusive and supportive virtual communities.}

\section{Conflict of Interest}
FS, PB were employed by Activision-Blizzard-King. The research was sponsored by \rev{Activision-Blizzard-King} under a sponsored research grant. The remaining authors declare that the research was conducted in the absence of any commercial or financial relationships that could be construed as a potential conflict of interest.


\bibliographystyle{ACM-Reference-Format}
\bibliography{sample-base, references}


\begin{thebibliography}{74}


\ifx \showCODEN    \undefined \def \showCODEN     #1{\unskip}     \fi
\ifx \showDOI      \undefined \def \showDOI       #1{#1}\fi
\ifx \showISBNx    \undefined \def \showISBNx     #1{\unskip}     \fi
\ifx \showISBNxiii \undefined \def \showISBNxiii  #1{\unskip}     \fi
\ifx \showISSN     \undefined \def \showISSN      #1{\unskip}     \fi
\ifx \showLCCN     \undefined \def \showLCCN      #1{\unskip}     \fi
\ifx \shownote     \undefined \def \shownote      #1{#1}          \fi
\ifx \showarticletitle \undefined \def \showarticletitle #1{#1}   \fi
\ifx \showURL      \undefined \def \showURL       {\relax}        \fi
\providecommand\bibfield[2]{#2}
\providecommand\bibinfo[2]{#2}
\providecommand\natexlab[1]{#1}
\providecommand\showeprint[2][]{arXiv:#2}

\bibitem[Achiam et~al\mbox{.}(2023)]%
        {achiam2023gpt}
\bibfield{author}{\bibinfo{person}{Josh Achiam}, \bibinfo{person}{Steven Adler}, \bibinfo{person}{Sandhini Agarwal}, \bibinfo{person}{Lama Ahmad}, \bibinfo{person}{Ilge Akkaya}, \bibinfo{person}{Florencia~Leoni Aleman}, \bibinfo{person}{Diogo Almeida}, \bibinfo{person}{Janko Altenschmidt}, \bibinfo{person}{Sam Altman}, \bibinfo{person}{Shyamal Anadkat}, {et~al\mbox{.}}} \bibinfo{year}{2023}\natexlab{}.
\newblock \showarticletitle{Gpt-4 technical report}.
\newblock \bibinfo{journal}{\emph{arXiv preprint arXiv:2303.08774}} (\bibinfo{year}{2023}).
\newblock


\bibitem[Bamler et~al\mbox{.}(2020)]%
        {bamler2020augmenting}
\bibfield{author}{\bibinfo{person}{Robert Bamler}, \bibinfo{person}{Farnood Salehi}, {and} \bibinfo{person}{Stephan Mandt}.} \bibinfo{year}{2020}\natexlab{}.
\newblock \showarticletitle{Augmenting and tuning knowledge graph embeddings}. In \bibinfo{booktitle}{\emph{Uncertainty in Artificial Intelligence}}. PMLR, \bibinfo{pages}{508--518}.
\newblock


\bibitem[Baumsteiger and Siegel(2018)]%
        {articlebau}
\bibfield{author}{\bibinfo{person}{Rachel Baumsteiger} {and} \bibinfo{person}{Jason Siegel}.} \bibinfo{year}{2018}\natexlab{}.
\newblock \showarticletitle{Measuring Prosociality: The Development of a Prosocial Behavioral Intentions Scale}.
\newblock \bibinfo{journal}{\emph{Journal of Personality Assessment}}  \bibinfo{volume}{101} (\bibinfo{date}{02} \bibinfo{year}{2018}).
\newblock
\urldef\tempurl%
\url{https://doi.org/10.1080/00223891.2017.1411918}
\showDOI{\tempurl}


\bibitem[Beres et~al\mbox{.}(2021)]%
        {beres2021don}
\bibfield{author}{\bibinfo{person}{Nicole~A Beres}, \bibinfo{person}{Julian Frommel}, \bibinfo{person}{Elizabeth Reid}, \bibinfo{person}{Regan~L Mandryk}, {and} \bibinfo{person}{Madison Klarkowski}.} \bibinfo{year}{2021}\natexlab{}.
\newblock \showarticletitle{Don’t you know that you’re toxic: Normalization of toxicity in online gaming}. In \bibinfo{booktitle}{\emph{Proceedings of the 2021 CHI conference on human factors in computing systems}}. \bibinfo{pages}{1--15}.
\newblock


\bibitem[Bibi(2024)]%
        {bibi2024legal}
\bibfield{author}{\bibinfo{person}{Palwasha Bibi}.} \bibinfo{year}{2024}\natexlab{}.
\newblock \showarticletitle{Legal and Ethical Challenges of Content Moderation: Balancing Privacy and Free Speech in the AI Era}.
\newblock  (\bibinfo{year}{2024}).
\newblock


\bibitem[Blei et~al\mbox{.}(2003)]%
        {10.5555/944919.944937}
\bibfield{author}{\bibinfo{person}{David~M. Blei}, \bibinfo{person}{Andrew~Y. Ng}, {and} \bibinfo{person}{Michael~I. Jordan}.} \bibinfo{year}{2003}\natexlab{}.
\newblock \showarticletitle{Latent Dirichlet Allocation}.
\newblock \bibinfo{journal}{\emph{J. Mach. Learn. Res.}} \bibinfo{volume}{3}, \bibinfo{number}{null} (\bibinfo{date}{mar} \bibinfo{year}{2003}), \bibinfo{pages}{993–1022}.
\newblock
\showISSN{1532-4435}


\bibitem[Bodenstedt et~al\mbox{.}(2019)]%
        {bodenstedt2019active}
\bibfield{author}{\bibinfo{person}{Sebastian Bodenstedt}, \bibinfo{person}{Dominik Rivoir}, \bibinfo{person}{Alexander Jenke}, \bibinfo{person}{Martin Wagner}, \bibinfo{person}{Michael Breucha}, \bibinfo{person}{Beat M{\"u}ller-Stich}, \bibinfo{person}{S{\"o}ren~Torge Mees}, \bibinfo{person}{J{\"u}rgen Weitz}, {and} \bibinfo{person}{Stefanie Speidel}.} \bibinfo{year}{2019}\natexlab{}.
\newblock \showarticletitle{Active learning using deep Bayesian networks for surgical workflow analysis}.
\newblock \bibinfo{journal}{\emph{International journal of computer assisted radiology and surgery}}  \bibinfo{volume}{14} (\bibinfo{year}{2019}), \bibinfo{pages}{1079--1087}.
\newblock


\bibitem[Chen et~al\mbox{.}(2018)]%
        {chen2018using}
\bibfield{author}{\bibinfo{person}{Nan-Chen Chen}, \bibinfo{person}{Margaret Drouhard}, \bibinfo{person}{Rafal Kocielnik}, \bibinfo{person}{Jina Suh}, {and} \bibinfo{person}{Cecilia~R Aragon}.} \bibinfo{year}{2018}\natexlab{}.
\newblock \showarticletitle{Using machine learning to support qualitative coding in social science: Shifting the focus to ambiguity}.
\newblock \bibinfo{journal}{\emph{ACM Transactions on Interactive Intelligent Systems (TiiS)}} \bibinfo{volume}{8}, \bibinfo{number}{2} (\bibinfo{year}{2018}), \bibinfo{pages}{1--20}.
\newblock


\bibitem[Chen et~al\mbox{.}(2020)]%
        {chen2020recall}
\bibfield{author}{\bibinfo{person}{Sanyuan Chen}, \bibinfo{person}{Yutai Hou}, \bibinfo{person}{Yiming Cui}, \bibinfo{person}{Wanxiang Che}, \bibinfo{person}{Ting Liu}, {and} \bibinfo{person}{Xiangzhan Yu}.} \bibinfo{year}{2020}\natexlab{}.
\newblock \showarticletitle{Recall and Learn: Fine-tuning Deep Pretrained Language Models with Less Forgetting}. In \bibinfo{booktitle}{\emph{Proceedings of the 2020 Conference on Empirical Methods in Natural Language Processing (EMNLP)}}. \bibinfo{pages}{7870--7881}.
\newblock


\bibitem[Decety et~al\mbox{.}(2016)]%
        {emphathy}
\bibfield{author}{\bibinfo{person}{Jean Decety}, \bibinfo{person}{Inbal Bartal}, \bibinfo{person}{Florina Uzefovsky}, {and} \bibinfo{person}{Ariel Knafo-Noam}.} \bibinfo{year}{2016}\natexlab{}.
\newblock \showarticletitle{Empathy as a driver of prosocial behaviour: Highly conserved neurobehavioural mechanisms across species}.
\newblock \bibinfo{journal}{\emph{Philosophical Transactions of the Royal Society B: Biological Sciences}}  \bibinfo{volume}{371} (\bibinfo{date}{01} \bibinfo{year}{2016}), \bibinfo{pages}{20150077}.
\newblock
\urldef\tempurl%
\url{https://doi.org/10.1098/rstb.2015.0077}
\showDOI{\tempurl}


\bibitem[Deterding and Waters(2021)]%
        {deterding2021flexible}
\bibfield{author}{\bibinfo{person}{Nicole~M Deterding} {and} \bibinfo{person}{Mary~C Waters}.} \bibinfo{year}{2021}\natexlab{}.
\newblock \showarticletitle{Flexible coding of in-depth interviews: A twenty-first-century approach}.
\newblock \bibinfo{journal}{\emph{Sociological methods \& research}} \bibinfo{volume}{50}, \bibinfo{number}{2} (\bibinfo{year}{2021}), \bibinfo{pages}{708--739}.
\newblock


\bibitem[Devlin et~al\mbox{.}(2018)]%
        {devlin2018bert}
\bibfield{author}{\bibinfo{person}{Jacob Devlin}, \bibinfo{person}{Ming-Wei Chang}, \bibinfo{person}{Kenton Lee}, {and} \bibinfo{person}{Kristina Toutanova}.} \bibinfo{year}{2018}\natexlab{}.
\newblock \showarticletitle{Bert: Pre-training of deep bidirectional transformers for language understanding}.
\newblock \bibinfo{journal}{\emph{arXiv preprint arXiv:1810.04805}} (\bibinfo{year}{2018}).
\newblock


\bibitem[Eder et~al\mbox{.}(2020)]%
        {eder2020anchor}
\bibfield{author}{\bibinfo{person}{Tobias Eder}, \bibinfo{person}{Viktor Hangya}, {and} \bibinfo{person}{Alexander Fraser}.} \bibinfo{year}{2020}\natexlab{}.
\newblock \showarticletitle{Anchor-based bilingual word embeddings for low-resource languages}.
\newblock \bibinfo{journal}{\emph{arXiv preprint arXiv:2010.12627}} (\bibinfo{year}{2020}).
\newblock


\bibitem[Eisenberg et~al\mbox{.}(2006)]%
        {articleeis}
\bibfield{author}{\bibinfo{person}{Nancy Eisenberg}, \bibinfo{person}{R.A. Fabes}, {and} \bibinfo{person}{Tracy Spinrad}.} \bibinfo{year}{2006}\natexlab{}.
\newblock \showarticletitle{Prosocial development. In N. Eisenberg, W. Damon, \& R. M. Lerner (Eds.),: Vol 3, Social, emotional, and personality development}.
\newblock \bibinfo{journal}{\emph{Handbook of child psychology}} (\bibinfo{date}{01} \bibinfo{year}{2006}), \bibinfo{pages}{646--718}.
\newblock


\bibitem[Elbagir and Yang(2019)]%
        {elbagir2019twitter}
\bibfield{author}{\bibinfo{person}{Shihab Elbagir} {and} \bibinfo{person}{Jing Yang}.} \bibinfo{year}{2019}\natexlab{}.
\newblock \showarticletitle{Twitter sentiment analysis using natural language toolkit and VADER sentiment}.
\newblock \bibinfo{journal}{\emph{Proceedings of the international multiconference of engineers and computer scientists}} \bibinfo{volume}{122}, \bibinfo{number}{16} (\bibinfo{year}{2019}).
\newblock


\bibitem[Fang and Xie(2023)]%
        {fang2023cert}
\bibfield{author}{\bibinfo{person}{Hongchao Fang} {and} \bibinfo{person}{Pengtao Xie}.} \bibinfo{year}{2023}\natexlab{}.
\newblock \showarticletitle{CERT: Contrastive Self-supervised Learning for Language Understanding}.
\newblock \bibinfo{journal}{\emph{Authorea Preprints}} (\bibinfo{year}{2023}).
\newblock


\bibitem[Farr et~al\mbox{.}(2024)]%
        {farr2024llm}
\bibfield{author}{\bibinfo{person}{David Farr}, \bibinfo{person}{Iain Cruickshank}, \bibinfo{person}{Nico Manzonelli}, \bibinfo{person}{Nicholas Clark}, \bibinfo{person}{Kate Starbird}, {and} \bibinfo{person}{Jevin West}.} \bibinfo{year}{2024}\natexlab{}.
\newblock \showarticletitle{LLM Confidence Evaluation Measures in Zero-Shot CSS Classification}.
\newblock \bibinfo{journal}{\emph{arXiv preprint arXiv:2410.13047}} (\bibinfo{year}{2024}).
\newblock


\bibitem[Findlay et~al\mbox{.}(2006)]%
        {FINDLAY2006347}
\bibfield{author}{\bibinfo{person}{Leanne~C. Findlay}, \bibinfo{person}{Alberta Girardi}, {and} \bibinfo{person}{Robert~J. Coplan}.} \bibinfo{year}{2006}\natexlab{}.
\newblock \showarticletitle{Links between empathy, social behavior, and social understanding in early childhood}.
\newblock \bibinfo{journal}{\emph{Early Childhood Research Quarterly}} \bibinfo{volume}{21}, \bibinfo{number}{3} (\bibinfo{year}{2006}), \bibinfo{pages}{347--359}.
\newblock
\showISSN{0885-2006}
\urldef\tempurl%
\url{https://doi.org/10.1016/j.ecresq.2006.07.009}
\showDOI{\tempurl}


\bibitem[Firoz et~al\mbox{.}(2023)]%
        {firoz2023automated}
\bibfield{author}{\bibinfo{person}{Neda Firoz}, \bibinfo{person}{Olga~Grigorievna Beresteneva}, \bibinfo{person}{Aksyonov~Sergey Vladimirovich}, \bibinfo{person}{Mohammad~Sadman Tahsin}, {and} \bibinfo{person}{Faiza Tafannum}.} \bibinfo{year}{2023}\natexlab{}.
\newblock \showarticletitle{Automated text-based depression detection using hybrid ConvLSTM and Bi-LSTM model}. In \bibinfo{booktitle}{\emph{2023 Third International Conference on Artificial Intelligence and Smart Energy (ICAIS)}}. IEEE, \bibinfo{pages}{734--740}.
\newblock


\bibitem[Frommel and Mandryk(2022)]%
        {frommel2022daily}
\bibfield{author}{\bibinfo{person}{Julian Frommel} {and} \bibinfo{person}{Regan~L Mandryk}.} \bibinfo{year}{2022}\natexlab{}.
\newblock \showarticletitle{Daily quests or daily pests? The benefits and Pitfalls of engagement rewards in games}.
\newblock \bibinfo{journal}{\emph{Proceedings of the ACM on Human-Computer Interaction}} \bibinfo{volume}{6}, \bibinfo{number}{CHI PLAY} (\bibinfo{year}{2022}), \bibinfo{pages}{1--23}.
\newblock


\bibitem[Furianto and Simanjuntak(2023)]%
        {furianto2023gaming}
\bibfield{author}{\bibinfo{person}{Furianto Furianto} {and} \bibinfo{person}{Risa~Rumentha Simanjuntak}.} \bibinfo{year}{2023}\natexlab{}.
\newblock \showarticletitle{Gaming Language as a Language Variations in Digital Humanities}. In \bibinfo{booktitle}{\emph{E3S Web of Conferences}}, Vol.~\bibinfo{volume}{388}. EDP Sciences, \bibinfo{pages}{04010}.
\newblock


\bibitem[Gallegos et~al\mbox{.}(2024)]%
        {gallegos2024bias}
\bibfield{author}{\bibinfo{person}{Isabel~O Gallegos}, \bibinfo{person}{Ryan~A Rossi}, \bibinfo{person}{Joe Barrow}, \bibinfo{person}{Md~Mehrab Tanjim}, \bibinfo{person}{Sungchul Kim}, \bibinfo{person}{Franck Dernoncourt}, \bibinfo{person}{Tong Yu}, \bibinfo{person}{Ruiyi Zhang}, {and} \bibinfo{person}{Nesreen~K Ahmed}.} \bibinfo{year}{2024}\natexlab{}.
\newblock \showarticletitle{Bias and fairness in large language models: A survey}.
\newblock \bibinfo{journal}{\emph{Computational Linguistics}} (\bibinfo{year}{2024}), \bibinfo{pages}{1--79}.
\newblock


\bibitem[Gao et~al\mbox{.}(2021)]%
        {gao2021simcse}
\bibfield{author}{\bibinfo{person}{T Gao}, \bibinfo{person}{X Yao}, {and} \bibinfo{person}{Danqi Chen}.} \bibinfo{year}{2021}\natexlab{}.
\newblock \showarticletitle{SimCSE: Simple Contrastive Learning of Sentence Embeddings}. In \bibinfo{booktitle}{\emph{EMNLP 2021-2021 Conference on Empirical Methods in Natural Language Processing, Proceedings}}.
\newblock


\bibitem[Gentile et~al\mbox{.}(2009a)]%
        {gentile2009effects}
\bibfield{author}{\bibinfo{person}{Douglas~A Gentile}, \bibinfo{person}{Craig~A Anderson}, \bibinfo{person}{Shintaro Yukawa}, \bibinfo{person}{Nobuko Ihori}, \bibinfo{person}{Muniba Saleem}, \bibinfo{person}{Lim~Kam Ming}, \bibinfo{person}{Akiko Shibuya}, \bibinfo{person}{Albert~K Liau}, \bibinfo{person}{Angeline Khoo}, \bibinfo{person}{Brad~J Bushman}, {et~al\mbox{.}}} \bibinfo{year}{2009}\natexlab{a}.
\newblock \showarticletitle{The effects of prosocial video games on prosocial behaviors: International evidence from correlational, longitudinal, and experimental studies}.
\newblock \bibinfo{journal}{\emph{Personality and Social Psychology Bulletin}} \bibinfo{volume}{35}, \bibinfo{number}{6} (\bibinfo{year}{2009}), \bibinfo{pages}{752--763}.
\newblock


\bibitem[Gentile et~al\mbox{.}(2009b)]%
        {doi:10.1177/0146167209333045}
\bibfield{author}{\bibinfo{person}{Douglas~A. Gentile}, \bibinfo{person}{Craig~A. Anderson}, \bibinfo{person}{Shintaro Yukawa}, \bibinfo{person}{Nobuko Ihori}, \bibinfo{person}{Muniba Saleem}, \bibinfo{person}{Lim~Kam Ming}, \bibinfo{person}{Akiko Shibuya}, \bibinfo{person}{Albert~K. Liau}, \bibinfo{person}{Angeline Khoo}, \bibinfo{person}{Brad~J. Bushman}, \bibinfo{person}{L.~Rowell Huesmann}, {and} \bibinfo{person}{Akira Sakamoto}.} \bibinfo{year}{2009}\natexlab{b}.
\newblock \showarticletitle{The Effects of Prosocial Video Games on Prosocial Behaviors: International Evidence From Correlational, Longitudinal, and Experimental Studies}.
\newblock \bibinfo{journal}{\emph{Personality and Social Psychology Bulletin}} \bibinfo{volume}{35}, \bibinfo{number}{6} (\bibinfo{year}{2009}), \bibinfo{pages}{752--763}.
\newblock
\urldef\tempurl%
\url{https://doi.org/10.1177/0146167209333045}
\showDOI{\tempurl}
\showeprint{https://doi.org/10.1177/0146167209333045}
\newblock
\shownote{PMID: 19321812}.


\bibitem[Grootendorst(2022)]%
        {grootendorst2022bertopic}
\bibfield{author}{\bibinfo{person}{Maarten Grootendorst}.} \bibinfo{year}{2022}\natexlab{}.
\newblock \bibinfo{title}{BERTopic: Neural topic modeling with a class-based TF-IDF procedure}.
\newblock
\newblock
\showeprint[arxiv]{2203.05794}~[cs.CL]


\bibitem[Hamilton(1964)]%
        {HAMILTON19641}
\bibfield{author}{\bibinfo{person}{W.D. Hamilton}.} \bibinfo{year}{1964}\natexlab{}.
\newblock \showarticletitle{The genetical evolution of social behaviour. I}.
\newblock \bibinfo{journal}{\emph{Journal of Theoretical Biology}} \bibinfo{volume}{7}, \bibinfo{number}{1} (\bibinfo{year}{1964}), \bibinfo{pages}{1--16}.
\newblock
\showISSN{0022-5193}
\urldef\tempurl%
\url{https://doi.org/10.1016/0022-5193(64)90038-4}
\showDOI{\tempurl}


\bibitem[Han et~al\mbox{.}(2024)]%
        {han2024chatgpt}
\bibfield{author}{\bibinfo{person}{Pengrui Han}, \bibinfo{person}{Rafal Kocielnik}, \bibinfo{person}{Adhithya Saravanan}, \bibinfo{person}{Roy Jiang}, \bibinfo{person}{Or Sharir}, {and} \bibinfo{person}{Anima Anandkumar}.} \bibinfo{year}{2024}\natexlab{}.
\newblock \showarticletitle{ChatGPT Based Data Augmentation for Improved Parameter-Efficient Debiasing of LLMs}.
\newblock \bibinfo{journal}{\emph{arXiv preprint arXiv:2402.11764}} (\bibinfo{year}{2024}).
\newblock


\bibitem[Hanu and {Unitary team}(2020)]%
        {Detoxify}
\bibfield{author}{\bibinfo{person}{Laura Hanu} {and} \bibinfo{person}{{Unitary team}}.} \bibinfo{year}{2020}\natexlab{}.
\newblock \bibinfo{title}{Detoxify}.
\newblock \bibinfo{howpublished}{Github. https://github.com/unitaryai/detoxify}.
\newblock


\bibitem[He et~al\mbox{.}(2021)]%
        {DBLP:journals/corr/abs-2111-09543}
\bibfield{author}{\bibinfo{person}{Pengcheng He}, \bibinfo{person}{Jianfeng Gao}, {and} \bibinfo{person}{Weizhu Chen}.} \bibinfo{year}{2021}\natexlab{}.
\newblock \showarticletitle{DeBERTaV3: Improving DeBERTa using ELECTRA-Style Pre-Training with Gradient-Disentangled Embedding Sharing}.
\newblock \bibinfo{journal}{\emph{CoRR}}  \bibinfo{volume}{abs/2111.09543} (\bibinfo{year}{2021}).
\newblock
\showeprint[arXiv]{2111.09543}
\urldef\tempurl%
\url{https://arxiv.org/abs/2111.09543}
\showURL{%
\tempurl}


\bibitem[Hilvert-Bruce and Neill(2020)]%
        {HILVERTBRUCE2020303}
\bibfield{author}{\bibinfo{person}{Zorah Hilvert-Bruce} {and} \bibinfo{person}{James~T. Neill}.} \bibinfo{year}{2020}\natexlab{}.
\newblock \showarticletitle{I'm just trolling: The role of normative beliefs in aggressive behaviour in online gaming}.
\newblock \bibinfo{journal}{\emph{Computers in Human Behavior}}  \bibinfo{volume}{102} (\bibinfo{year}{2020}), \bibinfo{pages}{303--311}.
\newblock
\showISSN{0747-5632}
\urldef\tempurl%
\url{https://doi.org/10.1016/j.chb.2019.09.003}
\showDOI{\tempurl}


\bibitem[Houlsby et~al\mbox{.}(2019)]%
        {houlsby2019parameterefficient}
\bibfield{author}{\bibinfo{person}{Neil Houlsby}, \bibinfo{person}{Andrei Giurgiu}, \bibinfo{person}{Stanislaw Jastrzebski}, \bibinfo{person}{Bruna Morrone}, \bibinfo{person}{Quentin de Laroussilhe}, \bibinfo{person}{Andrea Gesmundo}, \bibinfo{person}{Mona Attariyan}, {and} \bibinfo{person}{Sylvain Gelly}.} \bibinfo{year}{2019}\natexlab{}.
\newblock \bibinfo{title}{Parameter-Efficient Transfer Learning for NLP}.
\newblock
\newblock
\showeprint[arxiv]{1902.00751}~[cs.LG]


\bibitem[Hu et~al\mbox{.}(2023)]%
        {hu2023llm}
\bibfield{author}{\bibinfo{person}{Zhiqiang Hu}, \bibinfo{person}{Lei Wang}, \bibinfo{person}{Yihuai Lan}, \bibinfo{person}{Wanyu Xu}, \bibinfo{person}{Ee-Peng Lim}, \bibinfo{person}{Lidong Bing}, \bibinfo{person}{Xing Xu}, \bibinfo{person}{Soujanya Poria}, {and} \bibinfo{person}{Roy Lee}.} \bibinfo{year}{2023}\natexlab{}.
\newblock \showarticletitle{LLM-Adapters: An Adapter Family for Parameter-Efficient Fine-Tuning of Large Language Models}. In \bibinfo{booktitle}{\emph{Proceedings of the 2023 Conference on Empirical Methods in Natural Language Processing}}. \bibinfo{pages}{5254--5276}.
\newblock


\bibitem[Huang et~al\mbox{.}(2023)]%
        {huang2023adasent}
\bibfield{author}{\bibinfo{person}{Yongxin Huang}, \bibinfo{person}{Kexin Wang}, \bibinfo{person}{Sourav Dutta}, \bibinfo{person}{Raj~Nath Patel}, \bibinfo{person}{Goran Glavaš}, {and} \bibinfo{person}{Iryna Gurevych}.} \bibinfo{year}{2023}\natexlab{}.
\newblock \bibinfo{title}{AdaSent: Efficient Domain-Adapted Sentence Embeddings for Few-Shot Classification}.
\newblock
\newblock
\showeprint[arxiv]{2311.00408}~[cs.CL]


\bibitem[HuggingFace(2024)]%
        {Trainer26:online}
\bibfield{author}{\bibinfo{person}{HuggingFace}.} \bibinfo{year}{2024}\natexlab{}.
\newblock \bibinfo{title}{Trainer}.
\newblock \bibinfo{howpublished}{\url{https://huggingface.co/docs/transformers/en/main_classes/trainer}}.
\newblock
\newblock
\shownote{(Accessed on 11/05/2024)}.


\bibitem[Hutto and Gilbert(2014)]%
        {Hutto_Gilbert_2014}
\bibfield{author}{\bibinfo{person}{C. Hutto} {and} \bibinfo{person}{Eric Gilbert}.} \bibinfo{year}{2014}\natexlab{}.
\newblock \showarticletitle{VADER: A Parsimonious Rule-Based Model for Sentiment Analysis of Social Media Text}.
\newblock \bibinfo{journal}{\emph{Proceedings of the International AAAI Conference on Web and Social Media}} \bibinfo{volume}{8}, \bibinfo{number}{1} (\bibinfo{date}{May} \bibinfo{year}{2014}), \bibinfo{pages}{216--225}.
\newblock
\urldef\tempurl%
\url{https://doi.org/10.1609/icwsm.v8i1.14550}
\showDOI{\tempurl}


\bibitem[Jiang et~al\mbox{.}(2023a)]%
        {jiang2023mistral}
\bibfield{author}{\bibinfo{person}{Albert~Q Jiang}, \bibinfo{person}{Alexandre Sablayrolles}, \bibinfo{person}{Arthur Mensch}, \bibinfo{person}{Chris Bamford}, \bibinfo{person}{Devendra~Singh Chaplot}, \bibinfo{person}{Diego de~las Casas}, \bibinfo{person}{Florian Bressand}, \bibinfo{person}{Gianna Lengyel}, \bibinfo{person}{Guillaume Lample}, \bibinfo{person}{Lucile Saulnier}, {et~al\mbox{.}}} \bibinfo{year}{2023}\natexlab{a}.
\newblock \showarticletitle{Mistral 7B}.
\newblock \bibinfo{journal}{\emph{arXiv preprint arXiv:2310.06825}} (\bibinfo{year}{2023}).
\newblock


\bibitem[Jiang et~al\mbox{.}(2023b)]%
        {jiang2023low}
\bibfield{author}{\bibinfo{person}{Zhiying Jiang}, \bibinfo{person}{Matthew Yang}, \bibinfo{person}{Mikhail Tsirlin}, \bibinfo{person}{Raphael Tang}, \bibinfo{person}{Yiqin Dai}, {and} \bibinfo{person}{Jimmy Lin}.} \bibinfo{year}{2023}\natexlab{b}.
\newblock \showarticletitle{“Low-resource” text classification: A parameter-free classification method with compressors}. In \bibinfo{booktitle}{\emph{Findings of the Association for Computational Linguistics: ACL 2023}}. \bibinfo{pages}{6810--6828}.
\newblock


\bibitem[Jin et~al\mbox{.}(2021)]%
        {jin2021transferability}
\bibfield{author}{\bibinfo{person}{Xisen Jin}, \bibinfo{person}{Francesco Barbieri}, \bibinfo{person}{Brendan Kennedy}, \bibinfo{person}{Aida~Mostafazadeh Davani}, \bibinfo{person}{Leonardo Neves}, {and} \bibinfo{person}{Xiang Ren}.} \bibinfo{year}{2021}\natexlab{}.
\newblock \showarticletitle{On Transferability of Bias Mitigation Effects in Language Model Fine-Tuning}. In \bibinfo{booktitle}{\emph{Proceedings of the 2021 Conference of the North American Chapter of the Association for Computational Linguistics: Human Language Technologies}}. \bibinfo{pages}{3770--3783}.
\newblock


\bibitem[Khosla et~al\mbox{.}(2020)]%
        {khosla2020supervised}
\bibfield{author}{\bibinfo{person}{Prannay Khosla}, \bibinfo{person}{Piotr Teterwak}, \bibinfo{person}{Chen Wang}, \bibinfo{person}{Aaron Sarna}, \bibinfo{person}{Yonglong Tian}, \bibinfo{person}{Phillip Isola}, \bibinfo{person}{Aaron Maschinot}, \bibinfo{person}{Ce Liu}, {and} \bibinfo{person}{Dilip Krishnan}.} \bibinfo{year}{2020}\natexlab{}.
\newblock \showarticletitle{Supervised contrastive learning}.
\newblock \bibinfo{journal}{\emph{Advances in neural information processing systems}}  \bibinfo{volume}{33} (\bibinfo{year}{2020}), \bibinfo{pages}{18661--18673}.
\newblock


\bibitem[Kocielnik et~al\mbox{.}(2023a)]%
        {kocielnik2023can}
\bibfield{author}{\bibinfo{person}{Rafal Kocielnik}, \bibinfo{person}{Sara Kangaslahti}, \bibinfo{person}{Shrimai Prabhumoye}, \bibinfo{person}{Meena Hari}, \bibinfo{person}{R.Michael Alvarez}, {and} \bibinfo{person}{Anima Anandkumar}.} \bibinfo{year}{2023}\natexlab{a}.
\newblock \showarticletitle{Can you label less by using out-of-domain data? Active \& transfer learning with few-shot instructions}. In \bibinfo{booktitle}{\emph{Transfer Learning for Natural Language Processing Workshop}}. PMLR, \bibinfo{pages}{22--32}.
\newblock


\bibitem[Kocielnik et~al\mbox{.}(2024)]%
        {kocielnik2024challenges}
\bibfield{author}{\bibinfo{person}{Rafal Kocielnik}, \bibinfo{person}{Zhuofang Li}, \bibinfo{person}{Claudia Kann}, \bibinfo{person}{Deshawn Sambrano}, \bibinfo{person}{Jacob Morrier}, \bibinfo{person}{Mitchell Linegar}, \bibinfo{person}{Carly Taylor}, \bibinfo{person}{Min Kim}, \bibinfo{person}{Nabiha Naqvie}, {and} \bibinfo{person}{Feri Soltani}.} \bibinfo{year}{2024}\natexlab{}.
\newblock \showarticletitle{Challenges in moderating disruptive player behavior in online competitive action games}.
\newblock \bibinfo{journal}{\emph{Frontiers in Computer Science}}  \bibinfo{volume}{6} (\bibinfo{year}{2024}), \bibinfo{pages}{1283735}.
\newblock


\bibitem[Kocielnik et~al\mbox{.}(2023b)]%
        {kocielnik2023biastestgpt}
\bibfield{author}{\bibinfo{person}{Rafal Kocielnik}, \bibinfo{person}{Shrimai Prabhumoye}, \bibinfo{person}{Vivian Zhang}, \bibinfo{person}{Roy Jiang}, \bibinfo{person}{R~Michael Alvarez}, {and} \bibinfo{person}{Anima Anandkumar}.} \bibinfo{year}{2023}\natexlab{b}.
\newblock \showarticletitle{Biastestgpt: Using chatgpt for social bias testing of language models}.
\newblock \bibinfo{journal}{\emph{arXiv preprint arXiv:2302.07371}} (\bibinfo{year}{2023}).
\newblock


\bibitem[Kuhn(2008)]%
        {Kuhn2008-KUHPD}
\bibfield{author}{\bibinfo{person}{Steven Kuhn}.} \bibinfo{year}{2008}\natexlab{}.
\newblock \showarticletitle{Prisoner's Dilemma}.
\newblock In \bibinfo{booktitle}{\emph{Stanford Encyclopedia of Philosophy}}, \bibfield{editor}{\bibinfo{person}{Ed~Zalta}} (Ed.). \bibinfo{publisher}{Policy Press}.
\newblock


\bibitem[Laparra et~al\mbox{.}(2021)]%
        {laparra2021review}
\bibfield{author}{\bibinfo{person}{Egoitz Laparra}, \bibinfo{person}{Aurelie Mascio}, \bibinfo{person}{Sumithra Velupillai}, {and} \bibinfo{person}{Timothy Miller}.} \bibinfo{year}{2021}\natexlab{}.
\newblock \showarticletitle{A review of recent work in transfer learning and domain adaptation for natural language processing of electronic health records}.
\newblock \bibinfo{journal}{\emph{Yearbook of medical informatics}} \bibinfo{volume}{30}, \bibinfo{number}{01} (\bibinfo{year}{2021}), \bibinfo{pages}{239--244}.
\newblock


\bibitem[Le et~al\mbox{.}(2024)]%
        {le2024impact}
\bibfield{author}{\bibinfo{person}{Thanh-Dung Le}, \bibinfo{person}{Ti~Ti Nguyen}, {and} \bibinfo{person}{Vu~Nguyen Ha}.} \bibinfo{year}{2024}\natexlab{}.
\newblock \showarticletitle{The Impact of LoRA Adapters for LLMs on Clinical NLP Classification Under Data Limitations}.
\newblock \bibinfo{journal}{\emph{arXiv preprint arXiv:2407.19299}} (\bibinfo{year}{2024}).
\newblock


\bibitem[Lees et~al\mbox{.}(2022)]%
        {lees2022new}
\bibfield{author}{\bibinfo{person}{Alyssa Lees}, \bibinfo{person}{Vinh~Q Tran}, \bibinfo{person}{Yi Tay}, \bibinfo{person}{Jeffrey Sorensen}, \bibinfo{person}{Jai Gupta}, \bibinfo{person}{Donald Metzler}, {and} \bibinfo{person}{Lucy Vasserman}.} \bibinfo{year}{2022}\natexlab{}.
\newblock \showarticletitle{A new generation of perspective api: Efficient multilingual character-level transformers}. In \bibinfo{booktitle}{\emph{Proceedings of the 28th ACM SIGKDD Conference on Knowledge Discovery and Data Mining}}. \bibinfo{pages}{3197--3207}.
\newblock


\bibitem[Leong et~al\mbox{.}(2023a)]%
        {leong-etal-2023-self}
\bibfield{author}{\bibinfo{person}{Chak Leong}, \bibinfo{person}{Yi Cheng}, \bibinfo{person}{Jiashuo Wang}, \bibinfo{person}{Jian Wang}, {and} \bibinfo{person}{Wenjie Li}.} \bibinfo{year}{2023}\natexlab{a}.
\newblock \showarticletitle{Self-Detoxifying Language Models via Toxification Reversal}. In \bibinfo{booktitle}{\emph{Proceedings of the 2023 Conference on Empirical Methods in Natural Language Processing}}, \bibfield{editor}{\bibinfo{person}{Houda Bouamor}, \bibinfo{person}{Juan Pino}, {and} \bibinfo{person}{Kalika Bali}} (Eds.). \bibinfo{publisher}{Association for Computational Linguistics}, \bibinfo{address}{Singapore}, \bibinfo{pages}{4433--4449}.
\newblock
\urldef\tempurl%
\url{https://doi.org/10.18653/v1/2023.emnlp-main.269}
\showDOI{\tempurl}


\bibitem[Leong et~al\mbox{.}(2023b)]%
        {leong2023self}
\bibfield{author}{\bibinfo{person}{Chak~Tou Leong}, \bibinfo{person}{Yi Cheng}, \bibinfo{person}{Jiashuo Wang}, \bibinfo{person}{Jian Wang}, {and} \bibinfo{person}{Wenjie Li}.} \bibinfo{year}{2023}\natexlab{b}.
\newblock \showarticletitle{Self-detoxifying language models via toxification reversal}.
\newblock \bibinfo{journal}{\emph{arXiv preprint arXiv:2310.09573}} (\bibinfo{year}{2023}).
\newblock


\bibitem[Litvack-Miller et~al\mbox{.}(1997)]%
        {LitvackMiller1997TheSO}
\bibfield{author}{\bibinfo{person}{Willa Litvack-Miller}, \bibinfo{person}{Daniel McDougall}, {and} \bibinfo{person}{David~M. Romney}.} \bibinfo{year}{1997}\natexlab{}.
\newblock \showarticletitle{The structure of empathy during middle childhood and its relationship to prosocial behavior.}
\newblock \bibinfo{journal}{\emph{Genetic, social, and general psychology monographs}}  \bibinfo{volume}{123 3} (\bibinfo{year}{1997}), \bibinfo{pages}{303--24}.
\newblock
\urldef\tempurl%
\url{https://api.semanticscholar.org/CorpusID:29982230}
\showURL{%
\tempurl}


\bibitem[Liu et~al\mbox{.}(2015)]%
        {liu2015short}
\bibfield{author}{\bibinfo{person}{Yanling Liu}, \bibinfo{person}{Zhaojun Teng}, \bibinfo{person}{Haiying Lan}, \bibinfo{person}{Xin Zhang}, {and} \bibinfo{person}{Dezhong Yao}.} \bibinfo{year}{2015}\natexlab{}.
\newblock \showarticletitle{Short-term effects of prosocial video games on aggression: an event-related potential study}.
\newblock \bibinfo{journal}{\emph{Frontiers in Behavioral Neuroscience}}  \bibinfo{volume}{9} (\bibinfo{year}{2015}), \bibinfo{pages}{193}.
\newblock


\bibitem[Makhijani et~al\mbox{.}(2021)]%
        {makhijani2021quest}
\bibfield{author}{\bibinfo{person}{Rahul Makhijani}, \bibinfo{person}{Parikshit Shah}, \bibinfo{person}{Vashist Avadhanula}, \bibinfo{person}{Caner Gocmen}, \bibinfo{person}{Nicol{\'a}s~E Stier-Moses}, {and} \bibinfo{person}{Juli{\'a}n Mestre}.} \bibinfo{year}{2021}\natexlab{}.
\newblock \showarticletitle{Quest: Queue simulation for content moderation at scale}.
\newblock \bibinfo{journal}{\emph{arXiv preprint arXiv:2103.16816}} (\bibinfo{year}{2021}).
\newblock


\bibitem[Margolis(1984)]%
        {margolis1984selfishness}
\bibfield{author}{\bibinfo{person}{Howard Margolis}.} \bibinfo{year}{1984}\natexlab{}.
\newblock \bibinfo{booktitle}{\emph{Selfishness, altruism, and rationality}}.
\newblock \bibinfo{publisher}{University of Chicago Press}.
\newblock


\bibitem[McHugh(2012)]%
        {mchugh2012interrater}
\bibfield{author}{\bibinfo{person}{Mary~L McHugh}.} \bibinfo{year}{2012}\natexlab{}.
\newblock \showarticletitle{Interrater reliability: the kappa statistic}.
\newblock \bibinfo{journal}{\emph{Biochemia medica}} \bibinfo{volume}{22}, \bibinfo{number}{3} (\bibinfo{year}{2012}), \bibinfo{pages}{276--282}.
\newblock


\bibitem[Nagle et~al\mbox{.}(2014)]%
        {zora101582}
\bibfield{author}{\bibinfo{person}{Aniket Nagle}, \bibinfo{person}{Peter Wolf}, \bibinfo{person}{Robert Riener}, {and} \bibinfo{person}{Domen Novak}.} \bibinfo{year}{2014}\natexlab{}.
\newblock \showarticletitle{The use of player-centered positive reinforcement to schedule in-game rewards inreases enjoyment and performance in a serious game}.
\newblock \bibinfo{journal}{\emph{International Journal of Serious Games}} \bibinfo{volume}{1}, \bibinfo{number}{4} (\bibinfo{year}{2014}), \bibinfo{pages}{35--47}.
\newblock
\showISSN{2384-8766}
\urldef\tempurl%
\url{https://doi.org/10.17083/ijsg.v1i4.47}
\showDOI{\tempurl}


\bibitem[Oriola and Kotz{\'e}(2020)]%
        {oriola2020improved}
\bibfield{author}{\bibinfo{person}{Oluwafemi Oriola} {and} \bibinfo{person}{Eduan Kotz{\'e}}.} \bibinfo{year}{2020}\natexlab{}.
\newblock \showarticletitle{Improved semi-supervised learning technique for automatic detection of South African abusive language on Twitter}.
\newblock \bibinfo{journal}{\emph{South African Computer Journal}} \bibinfo{volume}{32}, \bibinfo{number}{2} (\bibinfo{year}{2020}), \bibinfo{pages}{56--79}.
\newblock


\bibitem[Penner et~al\mbox{.}(2005)]%
        {penner2005prosocial}
\bibfield{author}{\bibinfo{person}{Louis~A Penner}, \bibinfo{person}{John~F Dovidio}, \bibinfo{person}{Jane~A Piliavin}, {and} \bibinfo{person}{David~A Schroeder}.} \bibinfo{year}{2005}\natexlab{}.
\newblock \showarticletitle{Prosocial behavior: Multilevel perspectives}.
\newblock \bibinfo{journal}{\emph{Annu. Rev. Psychol.}}  \bibinfo{volume}{56} (\bibinfo{year}{2005}), \bibinfo{pages}{365--392}.
\newblock


\bibitem[Poeller et~al\mbox{.}(2023)]%
        {poeller2023suspecting}
\bibfield{author}{\bibinfo{person}{Susanne Poeller}, \bibinfo{person}{Martin~Johannes Dechant}, \bibinfo{person}{Madison Klarkowski}, {and} \bibinfo{person}{Regan~L Mandryk}.} \bibinfo{year}{2023}\natexlab{}.
\newblock \showarticletitle{Suspecting sarcasm: how league of legends players dismiss positive communication in toxic environments}.
\newblock \bibinfo{journal}{\emph{Proceedings of the ACM on Human-Computer Interaction}} \bibinfo{volume}{7}, \bibinfo{number}{CHI PLAY} (\bibinfo{year}{2023}), \bibinfo{pages}{1--26}.
\newblock


\bibitem[Prabhumoye et~al\mbox{.}(2021)]%
        {prabhumoye2021few}
\bibfield{author}{\bibinfo{person}{Shrimai Prabhumoye}, \bibinfo{person}{Rafal Kocielnik}, \bibinfo{person}{Mohammad Shoeybi}, \bibinfo{person}{Anima Anandkumar}, {and} \bibinfo{person}{Bryan Catanzaro}.} \bibinfo{year}{2021}\natexlab{}.
\newblock \showarticletitle{Few-shot instruction prompts for pretrained language models to detect social biases}.
\newblock \bibinfo{journal}{\emph{arXiv preprint arXiv:2112.07868}} (\bibinfo{year}{2021}).
\newblock


\bibitem[Reddy(2022)]%
        {reddy2022benchmarking}
\bibfield{author}{\bibinfo{person}{Charan Reddy}.} \bibinfo{year}{2022}\natexlab{}.
\newblock \showarticletitle{Benchmarking bias mitigation algorithms in representation learning through fairness metrics}.
\newblock \bibinfo{journal}{\emph{Thirty-fifth Conference on Neural Information Processing Systems Datasets and Benchmarks Track}} (\bibinfo{year}{2022}).
\newblock


\bibitem[Reddy et~al\mbox{.}(2018)]%
        {reddy2018semi}
\bibfield{author}{\bibinfo{person}{YCAP Reddy}, \bibinfo{person}{P Viswanath}, {and} \bibinfo{person}{B~Eswara Reddy}.} \bibinfo{year}{2018}\natexlab{}.
\newblock \showarticletitle{Semi-supervised learning: A brief review}.
\newblock \bibinfo{journal}{\emph{Int. J. Eng. Technol}} \bibinfo{volume}{7}, \bibinfo{number}{1.8} (\bibinfo{year}{2018}), \bibinfo{pages}{81}.
\newblock


\bibitem[Reimers and Gurevych(2019a)]%
        {DBLP:journals/corr/abs-1908-10084}
\bibfield{author}{\bibinfo{person}{Nils Reimers} {and} \bibinfo{person}{Iryna Gurevych}.} \bibinfo{year}{2019}\natexlab{a}.
\newblock \showarticletitle{Sentence-BERT: Sentence Embeddings using Siamese BERT-Networks}.
\newblock \bibinfo{journal}{\emph{CoRR}}  \bibinfo{volume}{abs/1908.10084} (\bibinfo{year}{2019}).
\newblock
\showeprint[arXiv]{1908.10084}
\urldef\tempurl%
\url{http://arxiv.org/abs/1908.10084}
\showURL{%
\tempurl}


\bibitem[Reimers and Gurevych(2019b)]%
        {reimers2019sentence}
\bibfield{author}{\bibinfo{person}{Nils Reimers} {and} \bibinfo{person}{Iryna Gurevych}.} \bibinfo{year}{2019}\natexlab{b}.
\newblock \showarticletitle{Sentence-bert: Sentence embeddings using siamese bert-networks}.
\newblock \bibinfo{journal}{\emph{arXiv preprint arXiv:1908.10084}} (\bibinfo{year}{2019}).
\newblock


\bibitem[Schmidt et~al\mbox{.}(2023)]%
        {schmidt2023one}
\bibfield{author}{\bibinfo{person}{Fabian~David Schmidt}, \bibinfo{person}{Ivan Vuli{\'c}}, {and} \bibinfo{person}{Goran Glava{\v{s}}}.} \bibinfo{year}{2023}\natexlab{}.
\newblock \showarticletitle{One For All \& All For One: Bypassing Hyperparameter Tuning with Model Averaging for Cross-Lingual Transfer}. In \bibinfo{booktitle}{\emph{Findings of the Association for Computational Linguistics: EMNLP 2023}}. \bibinfo{pages}{12186--12193}.
\newblock


\bibitem[Seiffert-Brockmann et~al\mbox{.}(2018)]%
        {seiffert2018memes}
\bibfield{author}{\bibinfo{person}{Jens Seiffert-Brockmann}, \bibinfo{person}{Trevor Diehl}, {and} \bibinfo{person}{Leonhard Dobusch}.} \bibinfo{year}{2018}\natexlab{}.
\newblock \showarticletitle{Memes as games: The evolution of a digital discourse online}.
\newblock \bibinfo{journal}{\emph{New Media \& Society}} \bibinfo{volume}{20}, \bibinfo{number}{8} (\bibinfo{year}{2018}), \bibinfo{pages}{2862--2879}.
\newblock


\bibitem[Shashua and Hazan(2005)]%
        {10.1145/1102351.1102451}
\bibfield{author}{\bibinfo{person}{Amnon Shashua} {and} \bibinfo{person}{Tamir Hazan}.} \bibinfo{year}{2005}\natexlab{}.
\newblock \showarticletitle{Non-Negative Tensor Factorization with Applications to Statistics and Computer Vision}. In \bibinfo{booktitle}{\emph{Proceedings of the 22nd International Conference on Machine Learning}} (Bonn, Germany) \emph{(\bibinfo{series}{ICML '05})}. \bibinfo{publisher}{Association for Computing Machinery}, \bibinfo{address}{New York, NY, USA}, \bibinfo{pages}{792–799}.
\newblock
\showISBNx{1595931805}
\urldef\tempurl%
\url{https://doi.org/10.1145/1102351.1102451}
\showDOI{\tempurl}


\bibitem[Team et~al\mbox{.}(2023)]%
        {team2023gemini}
\bibfield{author}{\bibinfo{person}{Gemini Team}, \bibinfo{person}{Rohan Anil}, \bibinfo{person}{Sebastian Borgeaud}, \bibinfo{person}{Yonghui Wu}, \bibinfo{person}{Jean-Baptiste Alayrac}, \bibinfo{person}{Jiahui Yu}, \bibinfo{person}{Radu Soricut}, \bibinfo{person}{Johan Schalkwyk}, \bibinfo{person}{Andrew~M Dai}, \bibinfo{person}{Anja Hauth}, {et~al\mbox{.}}} \bibinfo{year}{2023}\natexlab{}.
\newblock \showarticletitle{Gemini: a family of highly capable multimodal models}.
\newblock \bibinfo{journal}{\emph{arXiv preprint arXiv:2312.11805}} (\bibinfo{year}{2023}).
\newblock


\bibitem[Tomkinson and Van Den~Ende(2022)]%
        {tomkinson2022thank}
\bibfield{author}{\bibinfo{person}{Sian Tomkinson} {and} \bibinfo{person}{Benn Van Den~Ende}.} \bibinfo{year}{2022}\natexlab{}.
\newblock \showarticletitle{‘thank you for your compliance’: Overwatch as a disciplinary system}.
\newblock \bibinfo{journal}{\emph{Games and culture}} \bibinfo{volume}{17}, \bibinfo{number}{2} (\bibinfo{year}{2022}), \bibinfo{pages}{198--218}.
\newblock


\bibitem[Trivers(1971)]%
        {Trivers1971-TRITEO-4}
\bibfield{author}{\bibinfo{person}{Robert~L. Trivers}.} \bibinfo{year}{1971}\natexlab{}.
\newblock \showarticletitle{The Evolution of Reciprocal Altruism}.
\newblock \bibinfo{journal}{\emph{Quarterly Review of Biology}} \bibinfo{volume}{46}, \bibinfo{number}{1} (\bibinfo{year}{1971}), \bibinfo{pages}{35--57}.
\newblock


\bibitem[Tunstall et~al\mbox{.}(2022)]%
        {tunstall2022efficient}
\bibfield{author}{\bibinfo{person}{Lewis Tunstall}, \bibinfo{person}{Nils Reimers}, \bibinfo{person}{Unso Eun~Seo Jo}, \bibinfo{person}{Luke Bates}, \bibinfo{person}{Daniel Korat}, \bibinfo{person}{Moshe Wasserblat}, {and} \bibinfo{person}{Oren Pereg}.} \bibinfo{year}{2022}\natexlab{}.
\newblock \bibinfo{title}{Efficient Few-Shot Learning Without Prompts}.
\newblock
\newblock
\showeprint[arxiv]{2209.11055}~[cs.CL]


\bibitem[Weld et~al\mbox{.}(2021)]%
        {weld2021conda}
\bibfield{author}{\bibinfo{person}{Henry Weld}, \bibinfo{person}{Guanghao Huang}, \bibinfo{person}{Jean Lee}, \bibinfo{person}{Tongshu Zhang}, \bibinfo{person}{Kunze Wang}, \bibinfo{person}{Xinghong Guo}, \bibinfo{person}{Siqu Long}, \bibinfo{person}{Josiah Poon}, {and} \bibinfo{person}{Caren Han}.} \bibinfo{year}{2021}\natexlab{}.
\newblock \showarticletitle{CONDA: a CONtextual Dual-Annotated dataset for in-game toxicity understanding and detection}. In \bibinfo{booktitle}{\emph{Findings of the Association for Computational Linguistics: ACL-IJCNLP 2021}}. \bibinfo{pages}{2406--2416}.
\newblock


\bibitem[Wijkstra et~al\mbox{.}(2023)]%
        {wijkstra2023help}
\bibfield{author}{\bibinfo{person}{Michel Wijkstra}, \bibinfo{person}{Katja Rogers}, \bibinfo{person}{Regan~L Mandryk}, \bibinfo{person}{Remco~C Veltkamp}, {and} \bibinfo{person}{Julian Frommel}.} \bibinfo{year}{2023}\natexlab{}.
\newblock \showarticletitle{Help, My Game Is Toxic! First Insights from a Systematic Literature Review on Intervention Systems for Toxic Behaviors in Online Video Games}. In \bibinfo{booktitle}{\emph{Companion Proceedings of the Annual Symposium on Computer-Human Interaction in Play}}. \bibinfo{pages}{3--9}.
\newblock


\bibitem[Xiao et~al\mbox{.}(2023)]%
        {xiao2023supporting}
\bibfield{author}{\bibinfo{person}{Ziang Xiao}, \bibinfo{person}{Xingdi Yuan}, \bibinfo{person}{Q~Vera Liao}, \bibinfo{person}{Rania Abdelghani}, {and} \bibinfo{person}{Pierre-Yves Oudeyer}.} \bibinfo{year}{2023}\natexlab{}.
\newblock \showarticletitle{Supporting qualitative analysis with large language models: Combining codebook with GPT-3 for deductive coding}. In \bibinfo{booktitle}{\emph{Companion proceedings of the 28th international conference on intelligent user interfaces}}. \bibinfo{pages}{75--78}.
\newblock


\bibitem[Yang et~al\mbox{.}(2023)]%
        {yang2023towards}
\bibfield{author}{\bibinfo{person}{Zachary Yang}, \bibinfo{person}{Nicolas Grenon-Godbout}, {and} \bibinfo{person}{Reihaneh Rabbany}.} \bibinfo{year}{2023}\natexlab{}.
\newblock \showarticletitle{Towards Detecting Contextual Real-Time Toxicity for In-Game Chat}. In \bibinfo{booktitle}{\emph{Findings of the Association for Computational Linguistics: EMNLP 2023}}. \bibinfo{pages}{9894--9906}.
\newblock


\end{thebibliography}

\appendix

\section{Prosocial Expert Knowledge}
\label{apx:prosocial_keywords}
\begin{figure*}[!htp]
  \includegraphics[width=1.0\textwidth]{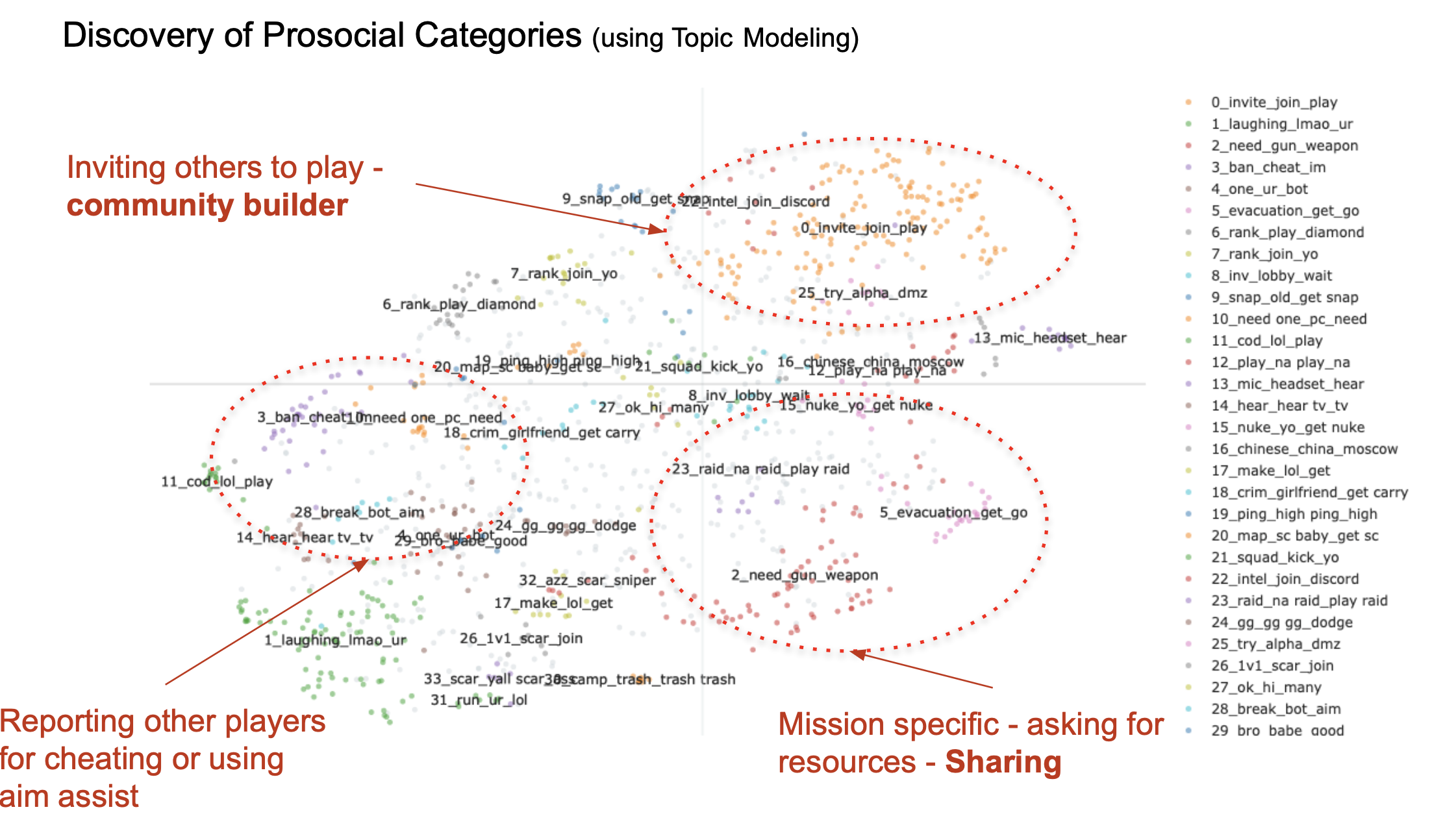}
  \caption{Topic Modeling Results and Cluster-keywords}
  \label{fig:topics}
\end{figure*}

We identify the following prosocial categories with keywords extracted from the best-aligned topics discovered using BERTopic framework \cite{grootendorst2022bertopic}, which are shown in Figure \ref{fig:topics}

\textbf{Sportsmanship} - ``GG'', ``Respect'', ``Trust'', ``Positive'', ``Honor'', ``Fair'', ``Gracious'', ``Proud'', "Cheers", "Props", "Thanks", "Good", "Kudos", "Team", "Play", "Well", "Humble", "Praise", "Clean", "Worthy"

\textbf{Sharing} - "Share", "Loot", "Ammo", "Med", "Ride", "Drop", "Gear", "Supply", "Resource", "Gift", "Pass", "Take", "Split", "Offer", "Grab", "Spot", "Help", "Save", "Cover", "Weapon"

\textbf{Good communicator} - "Ping", "Help", "Loot", "Enemy", "Callout", "Guide", "Spot", "Valuable", "Alert", "Watch", "Novice", "Mark", "Info", "Teach", "Close", "Assist", "Share", "Drop", "Cover", "Advise"

\textbf{Reciprocity} - "Revived", "Returned", "Back", "Favor", "Owe", "Trade", "Payback", "Thanks", "Even", "Rescue", "Matched", "Save", "Ready", "Turn", "Help", "Gotcha", "Owed", "Repay", "Trust", "Reciprocate"

\textbf{Community builder} - "Play", “Again”, "Respawn", "Land", "Party", "Join", "Stay", "Drop", "Team", "Invite", "Near", "Together", "Regroup", "Play", "Follow", "Spot", "Backup", "Help", "Guide", "Stick", "Support"

\newpage
\section{Non-English Language Counts}
\label{apx:non_english_counts}

\begin{table}[h!]
\caption{\rev{Language counts and percentages out of the 960 total instances in the labeled dataset.}}
\centering
\begin{tabular}{lcc}
\hline
\textbf{Language} & \textbf{Count} & \textbf{Percentage (\%)} \\
\hline
Spanish           & 10             & 1.04                     \\
Persian           & 7              & 0.73                     \\
Russian           & 6              & 0.62                     \\
Chinese           & 6              & 0.62                     \\
Filipino          & 6              & 0.62                     \\
Arabic            & 5              & 0.52                     \\
German            & 4              & 0.42                     \\
French            & 4              & 0.42                     \\
Portuguese        & 4              & 0.42                     \\
Indonesian        & 3              & 0.31                     \\
Hungarian         & 3              & 0.31                     \\
English           & 2              & 0.21                     \\
Czech             & 1              & 0.10                     \\
Sinhalese         & 1              & 0.10                     \\
Malay/Indonesian  & 1              & 0.10                     \\
Thai              & 1              & 0.10                     \\
Italian           & 1              & 0.10                     \\
Korean            & 1              & 0.10                     \\
Spanish, Turkish  & 1              & 0.10                     \\
Portuguese, Spanish & 1            & 0.10                     \\
Afrikaans         & 1              & 0.10                     \\
Polish            & 1              & 0.10                     \\
\hline
\end{tabular}
\label{tab-apx:languages}
\end{table}

\end{document}